\documentclass{article}
\usepackage{spconf,amsmath,graphicx}

\usepackage[colorlinks=true,allcolors=black]{hyperref}
\usepackage{url}
\usepackage{textcomp} 
\usepackage{gensymb} 
\usepackage{arydshln}
\usepackage{booktabs}
\usepackage{multirow}
\usepackage[utf8]{inputenc}
\usepackage[acronym]{glossaries}
\usepackage[caption=false,farskip=0pt]{subfig}
\usepackage[table,dvipsnames]{xcolor}
\usepackage{xspace} 
\usepackage[sort&compress,numbers]{natbib}
\usepackage{balance}

\usepackage{diagbox}

\newcommand\major[1]{{#1}}

\newacronymstyle{long-short-br}
{%
  \GlsUseAcrEntryDispStyle{long-short}%
}%
{%
  \GlsUseAcrStyleDefs{long-short}%
}
\setacronymstyle{long-short-br}

\title{\Large Towards Image-based Automatic Meter Reading in\\Unconstrained Scenarios: A Robust and Efficient Approach}

\makeatletter
\def\@name{ \emph{Rayson Laroca, Alessandra~B.~Araujo, Luiz~A.~Zanlorensi, Eduardo~C.~de~Almeida, David~Menotti}\thanks{
This is an author-prepared version. The published version is available at the \textit{IEEE Xplore Digital Library} (DOI: \href{http://doi.org/10.1109/ACCESS.2021.3077415}{\textcolor{blue}{10.1109/ACCESS.2021.3077415}}).} \\}
\makeatother

\address{\\[-10pt]Department of Informatics, Federal University of Paran\'a, Curitiba, Brazil\\[0.5ex] \textit{\{rblsantos, abaraujo, lazjunior, eduardo, menotti\}@inf.ufpr.br}}

\begin{document}
\sloppy
\maketitle
\newacronym{amr}{AMR}{Automatic Meter Reading}
\newacronym{bflop}{BFLOP}{billion floating-point operations}
\newacronym{cdcc}{CDCC-NET}{Corner Detection and Counter Classification Network}
\newacronym{copel}{Copel}{Energy Company of Paran\'a}
\newacronym{cnn}{CNN}{Convolutional Neural Network}
\newacronym{ctc}{CTC}{Connectionist Temporal Classification}
\newacronym{fcn}{FCN}{Fully Convolutional Network}
\newacronym{flop}{FLOP}{floating-point operations}
\newacronym{fps}{FPS}{frames per second}
\newacronym{gpu}{GPU}{Graphics Processing Unit}
\newacronym{iou}{IoU}{Intersection over Union}
\newacronym{map}{mAP}{mean Average Precision}
\newacronym{mser}{MSER}{Maximally Stable Extremal Regions}
\newacronym{ocr}{OCR}{Optical Character Recognition}
\newacronym{relu}{ReLU}{Rectified Linear Unit}
\newacronym{roi}{ROI}{region of interest}
\newacronym{rnn}{RNN}{Recurrent Neural Network}
\newacronym{sgd}{SGD}{Stochastic Gradient Descent}
\newacronym{ssd}{SSD}{Single Shot MultiBox Detector}

\newcommand{\scut}{SCUT-WMN\xspace}
\newcommand{\dataset}{Copel-AMR\xspace}
\newcommand{\ufpramr}{UFPR-AMR\xspace}

\newcommand{\cdcc}{Corner Detection and Counter Classification\xspace}
\newcommand{\crnet}{CR-NET\xspace}
\newcommand{\faster}{Faster R-CNN\xspace}

\newcommand{\detnet}{\major{Fast-YOLOv4-SmallObj}\xspace}
\newcommand{\ocrnet}{Fast-OCR\xspace}

\newcommand{\locatenet}{LocateNet\xspace}
\newcommand{\smallerlocatenet}{Smaller-LocateNet\xspace}
\newcommand{\yoohybrid}{Hybrid-MobileNetV2\xspace}

\newcommand{\retinanet}{RetinaNet\xspace}
\newcommand{\numtotal}{12{,}500\xspace}
\newcommand{\numok}{10{,}000\xspace}
\newcommand{\numfaulty}{2{,}500\xspace}
\newcommand{\supplementary}{https://web.inf.ufpr.br/vri/publications/amr-unconstrained-scenarios/}

\newcommand{\msperimage}{18\xspace}
\newcommand{\fps}{55\xspace}
\newcommand{\fpsmodified}{78\xspace}
\newcommand{\errorsavoided}{34\xspace}
\newcommand{\numbaselines}{10\xspace}
\newcommand{\numbaselinescdcc}{three\xspace}
\newcommand{\nruns}{10\xspace}

\newcommand{\fpscdcc}{191\xspace}

\newcommand{\minbaselines}{86.09}
\newcommand{\maxbaselines}{94.73}

\newcommand{\unrectaccufpr}{94.37}
\newcommand{\unrectaccdataset}{95.43}
\newcommand{\unrectaccaverage}{94.90}

\newcommand{\accufpr}{94.75}
\newcommand{\accdataset}{96.98}
\newcommand{\accaverage}{95.87}

\newcommand{\accaveragegt}{96.53}

\newcommand{\acclegible}{99.82}
\newcommand{\accfaulty}{98.9}

\begin{abstract}
\textit{Existing approaches for image-based \gls*{amr} have been evaluated on images captured in well-controlled scenarios.
However, real-world meter reading presents unconstrained  scenarios that are way more challenging due to dirt, various lighting conditions, scale variations, in-plane and out-of-plane rotations, among other factors.
In this work, we present an end-to-end approach for \gls*{amr} focusing on unconstrained scenarios.
Our main contribution is the insertion of a new stage in the \gls*{amr} pipeline, called \textit{corner detection and counter classification}, which enables the counter region to be rectified --~as well as the rejection of illegible/faulty~meters~-- prior to the recognition stage.
We also introduce a publicly available dataset, called \dataset, that contains \numtotal meter images acquired in the field by the service company's employees themselves, including \numfaulty images of faulty meters or cases where the reading is illegible due to occlusions.
Experimental evaluation demonstrates that the proposed system, which has three networks operating in a cascaded mode, outperforms all baselines in terms of recognition rate while still being quite efficient.
Moreover, as very few reading errors are tolerated in real-world applications, we show that our \gls*{amr} system achieves impressive recognition rates (i.e.,~$\ge$~99\%) when rejecting readings made with lower confidence~values.}
\end{abstract}

%
%
\section{Introduction}
\label{sec:introduction}

\glsresetall

\gls*{amr} refers to the technology whose goal is to automatically record the consumption of electric energy, gas and water for both monitoring and billing~\cite{khalifa2011survey,kabalci2016survey}.
Although smart meters are gradually replacing old meters, in many regions (especially in developing countries) the reading~is still done manually in the field, on a monthly basis, by an employee of the service company who takes a picture as reading proof~\cite{laroca2019convolutional, marques2019image}.

As such a procedure is prone to errors~\cite{vanetti2013gas,gallo2015robust,waqar2019meter}, the picture needs to be verified by another employee in some situations, for example, when the consumer makes a complaint about the amount charged and when the registered consumption differs significantly from that consumer's average.
This offline checking is known to be a laborious task~\cite{cerman2016mobile,quintanilha2017automatic}.

In this context, image-based techniques for \gls*{amr} are much needed, especially taking into account that it is not feasible to quickly replace old meters with smart ones~\cite{koscevic2018automatic,li2019light,tsai2019use}.
The idea behind image-based \gls*{amr}\major{, which is an specific scenario of scene text detection and recognition,} is that the aforementioned inspection can be carried out automatically, reducing mistakes introduced by the human factor and saving manpower~\cite{laroca2019convolutional,zuo2020robust}.
As pointed out by Salomon et al.~\cite{salomon2020deep}, the consumers themselves can capture photos of meters using a mobile device (e.g., a cell phone or a tablet). 
This eliminates the need for employees of the service company traveling around to perform local meter reading at each consumer unit, resulting in cost savings (especially in rural~areas).

Although \gls*{amr} (hereinafter \gls*{amr} refers to image-based \gls*{amr}) has received great attention in recent years, most works in the literature are still limited in several ways.
In general, the experiments were performed either on proprietary datasets~\cite{gomez2018cutting,koscevic2018automatic,marques2019image} or on datasets containing images captured on well-controlled environments~\cite{laroca2019convolutional,li2019light,kanagarathinam2019text}.
This is in stark contrast to related research areas, such as automatic license plate recognition, where in recent years the research focus shifted to unconstrained scenarios (with challenging factors such as blur, various lighting conditions, scale variations, in-plane and out-of-plane rotations, occlusions, etc.)~\cite{silva2018license}, helping to advance the state of the art considerably.
In addition, there are many works focused on a single stage of the \gls*{amr} pipeline~\cite{tsai2019use,waqar2019meter,yang2019fully}, which makes it difficult to accurately evaluate the presented methods in an end-to-end manner (e.g., the results achieved by a recognition model may vary considerably depending on how accurate the counter region is detected).
Another factor that makes it difficult to assess existing methods, or their applicability, is that the authors commonly do not report the execution time of the proposed approaches or the hardware in which they performed their experiments~\cite{kanagarathinam2019text,li2019light,yang2019fully}.
Finally, to the best of our knowledge, no previous work dealt with cases where it is not possible to perform the meter reading due to occlusions or faulty~meters, even though such cases are relatively common in~practice.

Considering the above discussion, in this work we present a novel end-to-end approach for \gls*{amr} that leverages the high capability of \glspl*{cnn} to achieve impressive results on real-world scenarios while still being quite efficient~--~it is capable of processing $\fps$~\gls*{fps} on a high-end \acrshort*{gpu}.
For our system to be both robust and efficient, we focused on achieving the best speed/accuracy trade-off at each stage when designing~it.
Our main contribution is the insertion of a new stage in the \gls*{amr} pipeline, called \textit{corner detection and counter classification}, where a multi-task network detects the four corners of the counter and simultaneously classifies it as legible/operational or illegible/faulty.
Prior to the recognition stage, legible counters are rectified using the predicted positions of the corners, thus improving the results obtained in distorted/inclined counters due to oblique views, and illegible counters are rejected.
We remark that while improving the recognition performance has an important role in reducing manual intervention, automatically classifying and filtering out illegible meter readings is of paramount importance to the service company, as such cases still require human~review.

As part of this work, we introduce a publicly available dataset\footnote{The \dataset dataset is publicly available to the research community at \url{https://web.inf.ufpr.br/vri/databases/copel-amr/}. Access is granted \textbf{upon request}, i.e., interested parties must register by filling out a registration form and agreeing to the dataset's terms of use.}, called \dataset, that contains $\numtotal$ meter images acquired in the field by the service company's employees themselves (i.e., the images were taken in real-world conditions), including $\numfaulty$ images of faulty meters or cases where the reading is illegible.
\major{More specifically, we consider as \textit{faulty} the meters where it is not possible to perform the meter reading because no reading is displayed (e.g., electronic meters where the display screen is blank) and as \textit{illegible} the meter images where it is not possible to perform the meter reading due to occlusions in the counter region (e.g., dirt, reflections, and/or water vapor on the meter glass).}
To the best of our knowledge, this is the first public dataset for end-to-end \gls*{amr} captured ``in the wild'' and also the only one with images of illegible/faulty meters.
The proposed dataset has six times more images --~and contains a larger variety in different aspects~-- than the largest dataset found in the literature for the evaluation of end-to-end \gls*{amr}~methods.

We experimentally evaluate the proposed approach in two public datasets: \ufpramr~\cite{laroca2019convolutional} and \dataset (described in Section~\ref{sec:dataset}).
Our system achieves state-of-the-art results by outperforming \numbaselines deep learning-based baselines in both datasets \major{(we are not aware of any work in the AMR literature where so many methods were evaluated in the~experiments)}.
The importance of the \textit{corner detection and counter classification} stage is demonstrated, as our system made~$\errorsavoided$\% fewer reading errors \major{in the legible/operational meters} of the \dataset dataset --~where the images were captured in unconstrained~scenarios~-- when feeding rectified counters into the recognition~network.
Moreover, simultaneously to the prediction of the counter corners, our network is able to filter out most of the illegible/faulty meters~(i.e.,~$\accfaulty$\%), thereby reducing the overall cost of the proposed system since the counter rectification and recognition tasks are skipped in such~cases, while correctly accepting $\acclegible$\% of the legible/operational~meters.

In summary, our paper has three main contributions:
\begin{itemize}
    \item A robust and efficient approach for \gls*{amr} that achieves state-of-the-art results in two public datasets and that significantly reduces the number of images that are sent to human review by filtering out most images containing illegible/faulty meters.
    Our system explores three carefully designed and optimized networks, operating in a cascaded mode, to achieve the best trade-off between accuracy and~speed.
    \item A public dataset for end-to-end \gls*{amr} with $12{,}500$ fully-annotated images acquired on real-world scenarios by the service company's employees themselves, \major{being $10{,}000$ of them of legible/operational meters and $2{,}500$ of illegible/faulty meters}.
    The dataset contains a well-defined evaluation protocol to assist the development of new approaches for \gls*{amr} as well as the fair comparison among published~works;
    \item A comparative assessment of the proposed approach and \numbaselines baseline methods based on deep learning, unlike most works in the literature that reported only the results obtained by the proposed methods or compared them exclusively with traditional approaches --~often carrying out experiments exclusively on proprietary datasets. 
    It is observed that most of the reading errors made by our \gls*{amr} system occurred in challenging cases, where even humans can make mistakes, as one digit becomes very similar to another due to artifacts in the counter~region.
\end{itemize}

The remainder of this paper is organized as follows.
We review related works in Section~\ref{sec:related_work}. The \dataset dataset is introduced in Section~\ref{sec:dataset}.
In Section~\ref{sec:proposed}, we present the proposed approach in detail. 
The experiments carried out and the results achieved are described in Section~\ref{sec:experiments}.
Conclusions and future works are given in Section~\ref{sec:conclusions}.
\section{Related Work}
\label{sec:related_work}

\major{Image-based AMR is a specific application of scene text detection and recognition}~\cite{gao2018automatic,gomez2018cutting}\major{.
However, there are some fundamental differences for the general task of detecting and recognizing scene text that should be highlighted: (i)~in the AMR context, there is only one region of interest in each image (i.e., the counter region) and all digits are within it; (ii)~recognition networks for AMR need to learn $10$ classes (digits $0$ to $9$) while networks for general scene text recognition need to learn $36$ character classes ($26$~letters and $10$~digits; the number of classes can be even higher depending on the language); and (iii)~AMR presents an unusual challenge in Optical Character Recognition~(OCR): rotating digits in electromechanical meters.
Typically, rotating digits are a major cause of errors in such meters, even when robust approaches are employed for digit/counter recognition}~{\cite{gao2018automatic,laroca2019convolutional}.\major{
This last point further emphasizes the importance of the proposed dataset, as recognition models trained exclusively on images from general datasets for robust reading (e.g., ICDAR 2013}~\cite{karatzas2013icdar}) \major{are likely to fail in these~cases.}

Over the last decade, a number of methods have been proposed for image-based \gls*{amr}.
Prior to the widespread adoption of deep learning in computer vision, most approaches to this task explored image enhancement techniques and handcrafted features with a similar pipeline, i.e., (i)~counter detection followed by (ii)~digit segmentation and (iii)~digit recognition~\cite{vanetti2013gas,gallo2015robust}.
Most limitations of such methods may be attributed to the fact that handcrafted features are easily affected by noise and are generally not robust to images captured under unconstrained environments~\cite{li2019light,liao2019reading,waqar2019meter}.

In deep learning-based methods~\cite{gomez2018cutting,marques2019image,yang2019fully}, on the other hand, usually the entire counter region is fed into the recognition network and all digits are predicted simultaneously (instead of first segmenting and then recognizing each of them).
As major advances have been achieved in computer vision through deep learning~\cite{lecun2015deep}, in this section we review works that employed deep learning-based approaches in the \gls*{amr}~context.

We also focus on studies related to digit-based meters, even though there are some recent works that addressed the recognition of dial meters~\cite{he2019value,salomon2020deep,zuo2020robust}.
Such works usually explore the angle between the pointer and the dial to perform the reading.


Object detectors have been explored frequently to deal with counter detection.
For example, Ko{\v s}{\v c}evi{\' c} \& Suba{\v s}i{\'c}~\cite{koscevic2018automatic} employed \faster~\cite{ren2017faster} to detect counters and serial numbers on images of residential meters, whereas Tsai et al.~\cite{tsai2019use} applied and fine-tuned \gls*{ssd}~\cite{liu2016ssd} for counter detection in electricity meters.
In both studies, only proprietary datasets were used to evaluate the detectors.

Similarly, Laroca et al.~\cite{laroca2019convolutional} tackled counter detection using Fast-YOLOv2~\cite{redmon2017yolo9000}.
Considering that all counter regions were correctly detected in their experiments, the authors stated that very deep models are not necessary to handle this~task. For counter recognition, three \gls*{cnn}-based approaches were evaluated, with the CR-NET model~\cite{silva2017realtime} outperforming two segmentation-free models~\cite{shi2017endtoend,goncalves2019multitask} in terms of recognition rate.
It should be noted that the images they used were acquired in a warehouse of the service company by one of the authors; in other words, the images are not as unconstrained as those collected in the field by the service company's employees (e.g., there is no external lighting or occlusions caused by railings or~vegetation).

Rather than exploring object detectors, Calefati et al.~\cite{calefati2019reading} employed a \gls*{fcn} for semantic segmentation~\cite{long2015fully} to handle the detection stage.
Then, the counter region was aligned horizontally through the application of traditional image-processing techniques, such as contours extraction and mathematical morphology, in the segmentation mask.
Finally, a \gls*{cnn} model was employed to produce the meter reading from the aligned counter region.
Although their experiments were carried out on real-world images, only a cropped version of their dataset is available for the research community (as only the region containing the digits was kept in each image, it is not possible to use the released dataset for the evaluation of end-to-end methods).
In addition, the accuracy rates obtained in some digit positions were significantly lower than in others due to the low variability in such positions (their dataset is biased and so is their recognition model; this phenomenon was also observed by Laroca et al.~\cite{laroca2019convolutional}).
Such a limitation must be addressed before an \gls*{amr} solution can be used in~practice. 

Yang et al.~\cite{yang2019fully} combined an \gls*{fcn} and \gls*{ctc} without any intermediate recurrent connections for counter recognition in water meter images.
Their network achieved better recognition results than two baselines, showing that such a network is capable of learning contextual information and thus eliminating the need for recurring layers.
However, it is important to note that their experiments were carried out only on manually-cropped counter regions and that such a segmentation-free approach may not be as robust in cases where the region of interest (here, the counter) is not detected as precisely~\cite{goncalves2018realtime}.

Taking into account the importance of designing highly efficient methods in the \gls*{amr} context, Li et al.~\cite{li2019light} proposed a light-weight \gls*{cnn} for counter recognition that splices a certain number of $1\times1$ and $3\times3$ kernels to reduce the network parameters with little loss in the recognition rate.
The results reported by them are impressive considering the accuracy/speed trade-off obtained; nevertheless, their experiments were performed exclusively on a private dataset with well-controlled images quite similar to each other (i.e., the images were captured by a camera installed in the meter box and preprocessed manually by the authors; thus, they have no blur, scale variations, shadows, occlusions, significant rotations, among other challenging~factors). 

\begin{figure*}[!htb]
    \centering
    
    \resizebox{0.8\linewidth}{!}{
    \includegraphics[width=0.16\linewidth]{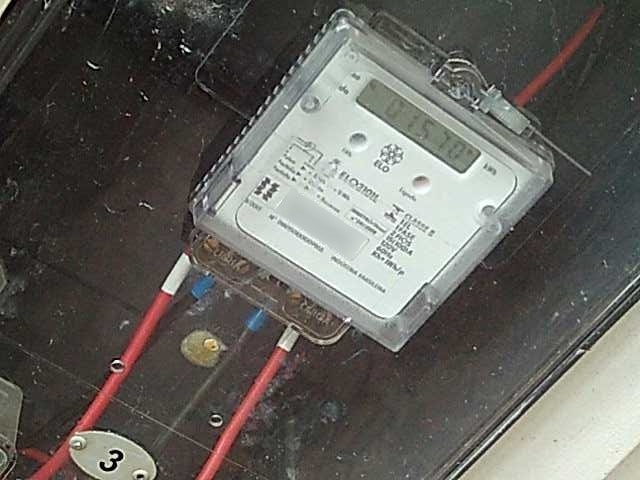}
    \includegraphics[width=0.16\linewidth]{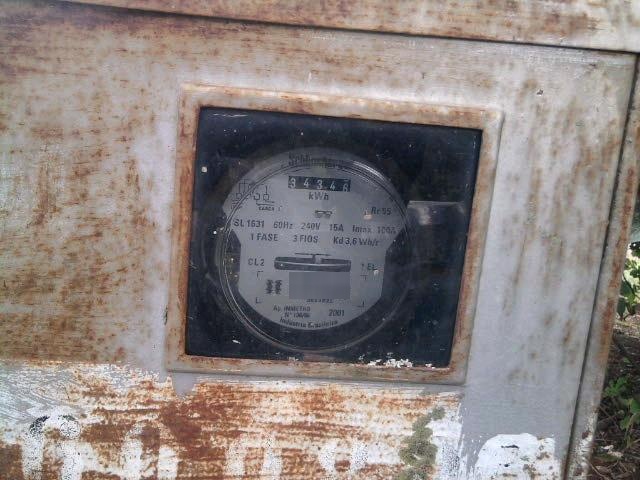}
    \includegraphics[width=0.16\linewidth]{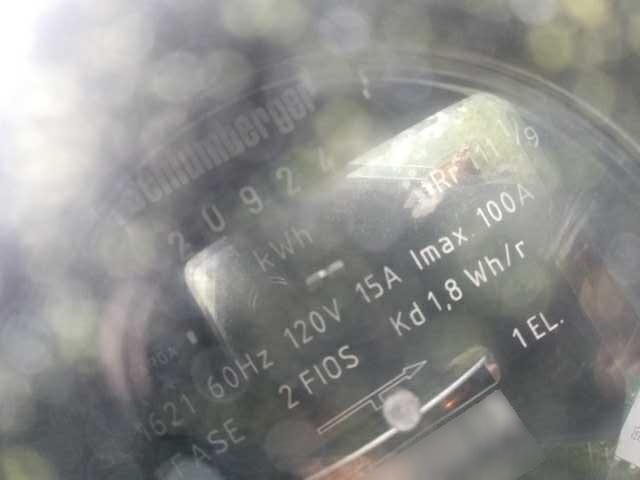}
    \includegraphics[width=0.16\linewidth]{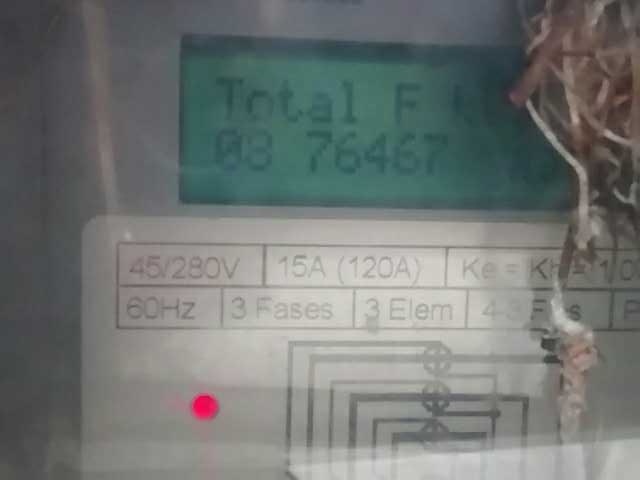}
    \includegraphics[width=0.16\linewidth]{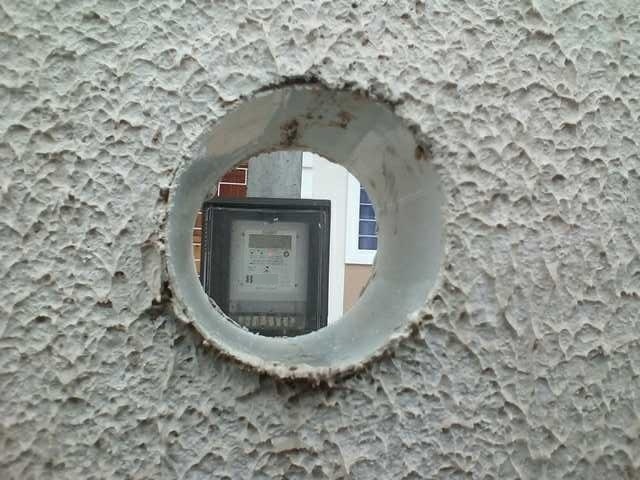}
    \includegraphics[width=0.16\linewidth]{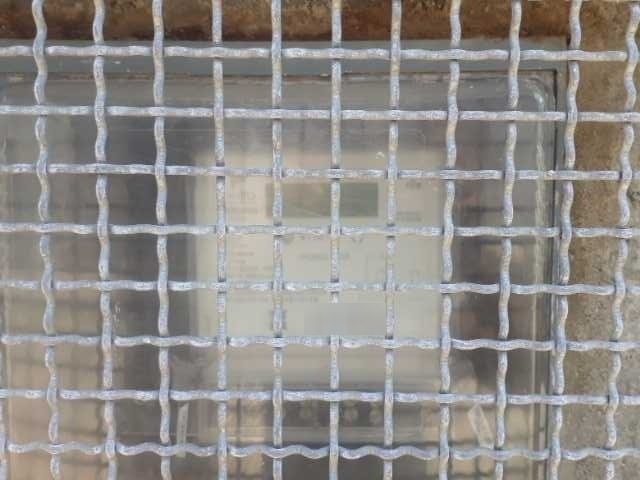}
    }
    
    \vspace{0.35mm}
    
    \resizebox{0.8\linewidth}{!}{
    \includegraphics[width=0.14\linewidth]{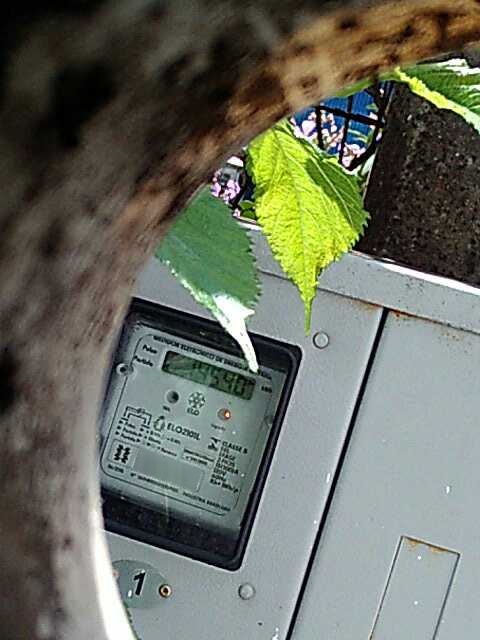}
    \includegraphics[width=0.14\linewidth]{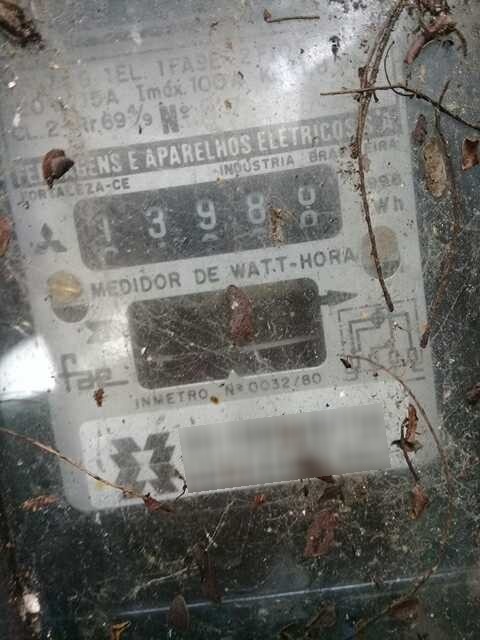}
    \includegraphics[width=0.14\linewidth]{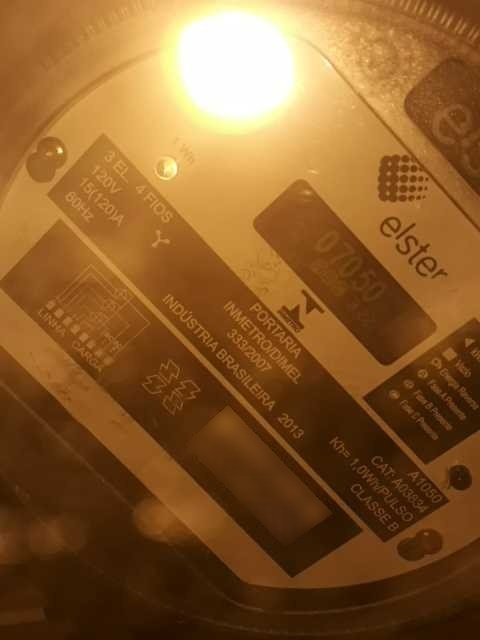}
    \includegraphics[width=0.14\linewidth]{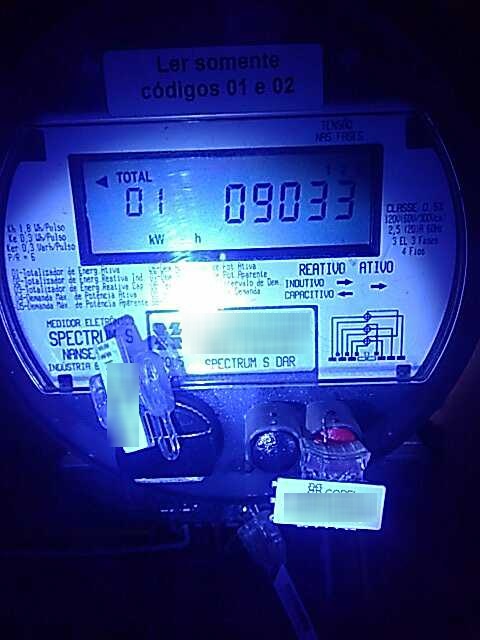}
    \includegraphics[width=0.14\linewidth]{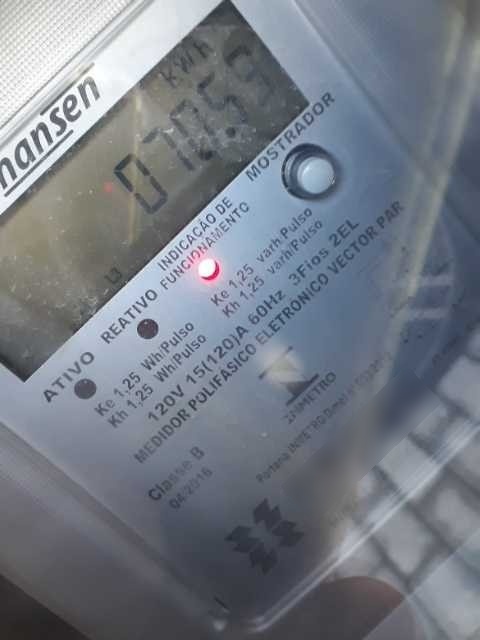}
    \includegraphics[width=0.14\linewidth]{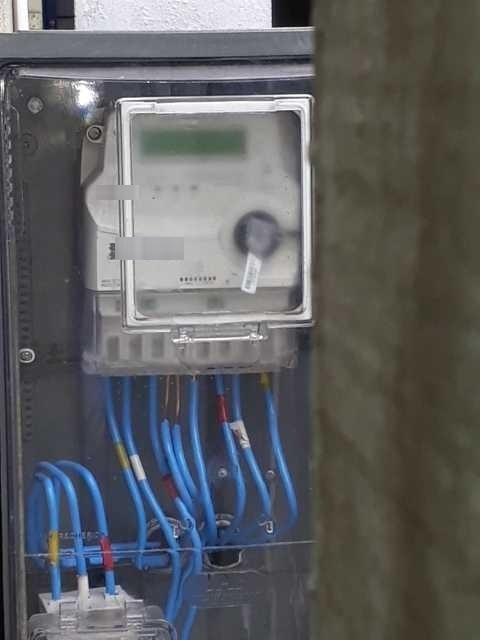}
    \includegraphics[width=0.14\linewidth]{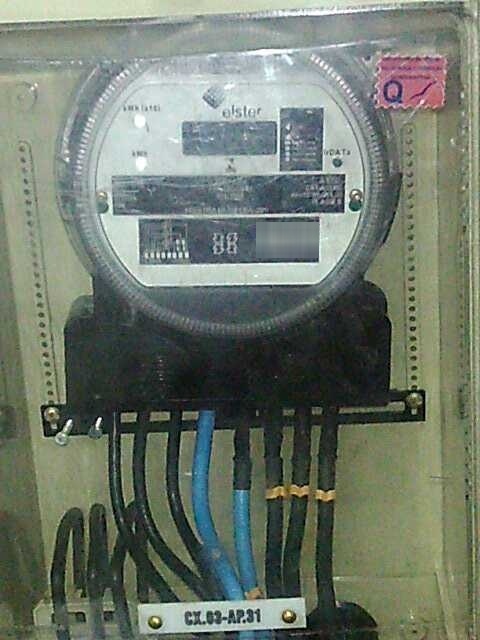}
    }
    
    \vspace{0.35mm}
    
    \resizebox{0.8\linewidth}{!}{
    \includegraphics[width=0.16\linewidth]{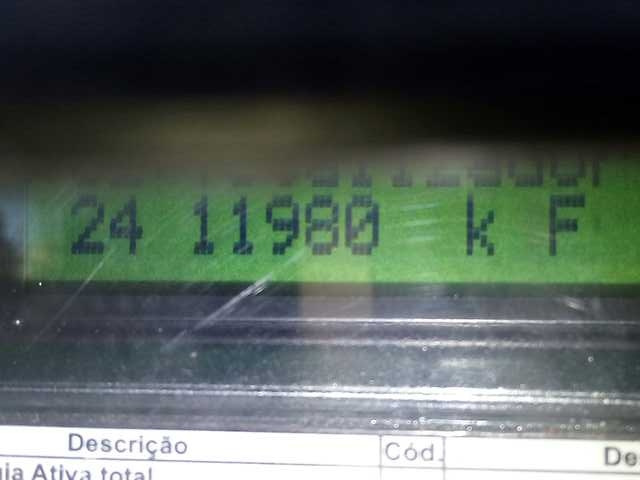}
    \includegraphics[width=0.16\linewidth]{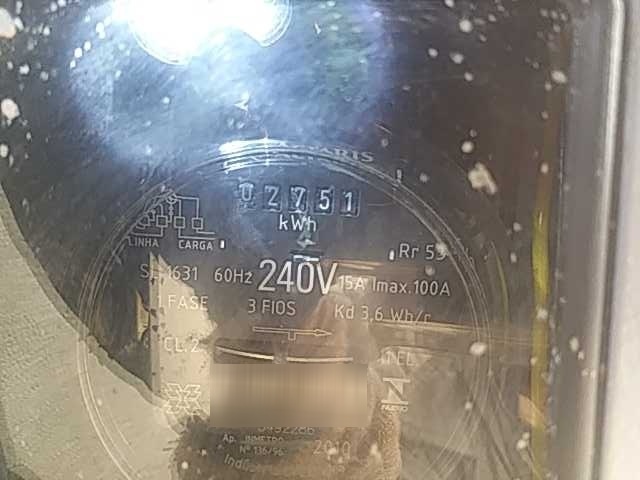}
    \includegraphics[width=0.16\linewidth]{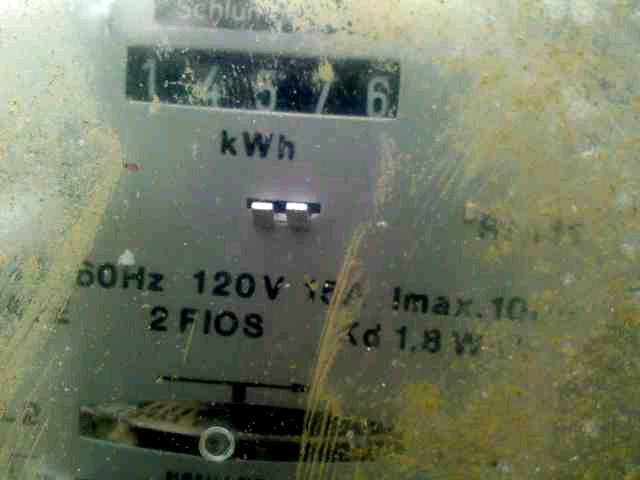}
    \includegraphics[width=0.16\linewidth]{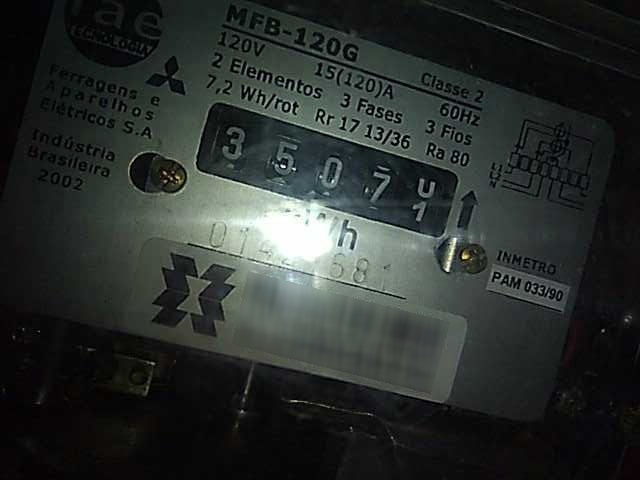}
    \includegraphics[width=0.16\linewidth]{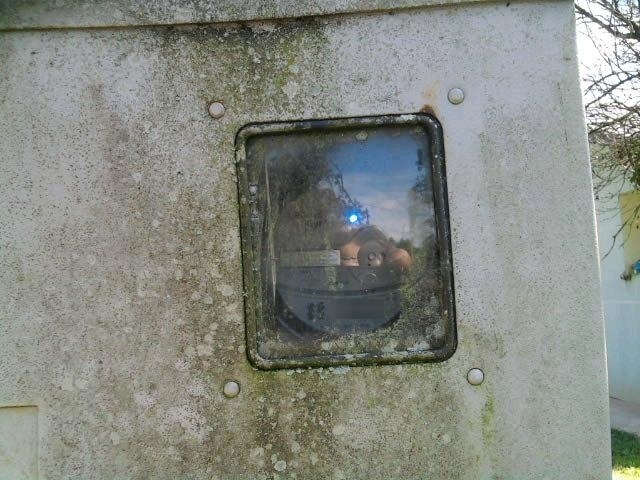}
    \includegraphics[width=0.16\linewidth]{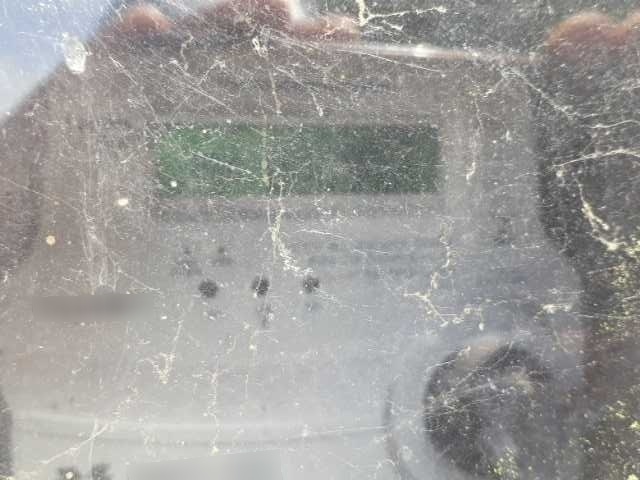}
    }
    
    \vspace{-1.5mm}
    
    \caption{\small Some images extracted from the \dataset dataset. 
    Note that there are both electromechanical and electronic meters \major{and that meters of different types/models often have different screen sizes and layouts}. 
    The last two images in each row are from faulty or illegible meters.
    As requested by Copel, the regions containing consumer identification were blurred on each image due to privacy~constraints.}
    \label{fig:samples-dataset}
\end{figure*}

Marques et al.~\cite{marques2019image} fine-tuned the \faster and \retinanet~\cite{lin2017focal} object detectors for counter recognition. 
Although the authors reported \gls*{map} rates above $90$\% with both detectors, only a small subset of counter images from a private dataset was employed in their experiments and the hardware used (i.e., the \acrshort*{gpu}) was not specified, making it difficult to compare  their methodology with previous works both in terms of efficiency and~recognition~rate.
Waqar et al.~\cite{waqar2019meter} also employed \faster for counter recognition, however, a low recognition rate of $76$\% was reported in their experiments. 
As in~\cite{calefati2019reading}, the accuracy achieved in some digit positions was considerably lower than in others, probably due to the fact that the authors did not take into account the bias in the distribution of the digit classes in the training set when fine-tuning the \faster model.
Despite the fact that the authors claimed that their method can be deployed in real-time applications, no experiments related to execution time were performed/reported.

There are some works in which the authors chose to perform the meter reading directly in the input image, i.e., without counter detection.
For instance, Liao et al.~\cite{liao2019reading} simply employed YOLOv3~\cite{redmon2018yolov3} for this task, whereas Gómez et al.~\cite{gomez2018cutting} proposed a \gls*{cnn} model that directly outputs the meter reading in a segmentation-free manner.
Although promising results were reported in these works, such approaches are not robust to severe perspective distortions and small-meter images~\cite{calefati2019reading, laroca2019convolutional}.

Considering the many limitations of existing works, we propose a novel end-to-end \gls*{amr} system that contains a unified approach for \textit{corner detection and counter classification} in order to (i)~improve the recognition results (especially in unconstrained scenarios) through counter rectification and (ii)~significantly reduce the number of images that are sent to human review by filtering out images containing illegible/faulty meters.
The proposed system is \major{empirically} evaluated in two public datasets that have well-defined evaluation protocols and that enable the evaluation of end-to-end \gls*{amr} methods.
One of the datasets, called \ufpramr~\cite{laroca2019convolutional}, has $4K$~images collected in a warehouse of the service company by one of its authors, i.e., under controlled capture conditions, while the other (introduced in Section~\ref{sec:dataset}) contains $480p$ images acquired in the field by the service company's employees themselves, i.e., under unconstrained capture environments.
In our experiments, detailed information regarding both the hardware/frameworks used and the execution time required to run our \gls*{amr} system is also provided in order to enable an accurate analysis of its speed/accuracy trade-off, as well as its~applicability. 
\section{The Copel-AMR dataset}
\label{sec:dataset}

The \dataset dataset contains $\numtotal$ meter images acquired in the field by the employees of the \gls*{copel}, which directly serves more than $4$ million consuming units, across $395$ cities and $1{,}113$ locations (i.e., districts, villages and settlements), located in the Brazilian state of Paraná~\cite{copel}.
Thus, \dataset is composed of images captured in unconstrained scenarios, which typically include blur (due to camera motion), dirt, scale variations, in-plane and out-of-plane rotations, reflections, shadows, and occlusions.
In $\numfaulty$ images (i.e., $20$\% of the dataset), it is not even possible to perform the meter reading due to occlusions or faulty meters. 
Although such situations are found on a daily basis by meter readers, there is no work in the literature addressing them or public datasets containing images of illegible/faulty meters, to the best of our knowledge.
Fig.~\ref{fig:samples-dataset} shows the diversity of the~dataset.
\major{Note that as the model of the meters being installed/replaced has changed over the years, there is a wide variety of meter types in our~dataset.}

The images have a resolution of $480\times640$ or $640\times480$ pixels, depending on the orientation in which they were taken.
Considering that the meter is operational and that there are no occlusions, these resolutions are enough for the meter reading to be legible.

For each image in our dataset, we manually labeled the meter reading, the position~($x$,~$y$) of each of the four corners of the counter, and a bounding box~($x$,~$y$, $w$,~$h$) for each digit.
Corner annotations --~which can be converted to a bounding box~-- enable the counter to be rectified, while bounding boxes enable the training of object~detectors as well as the application of a wider range of data augmentation techniques.
\major{As far as we are aware, the Copel-AMR dataset is the only one to provide so much labeled information for each~image.}

As the \dataset dataset contains $10{,}000$ images of legible/operational meters and each meter reading consists of $5$ digits, we manually labeled a total of $50{,}000$ digits.
The distribution of the digit classes in the dataset is shown in Fig.~\ref{fig:frequency}.
We observed that the digit `$0$' has many more instances than the others, which was expected, due to the fact that a brand new meter starts with \texttt{00000} and the leftmost digit positions take longer to be~increased.

\begin{figure}[!htb]
    \centering
    \includegraphics[width=0.85\linewidth]{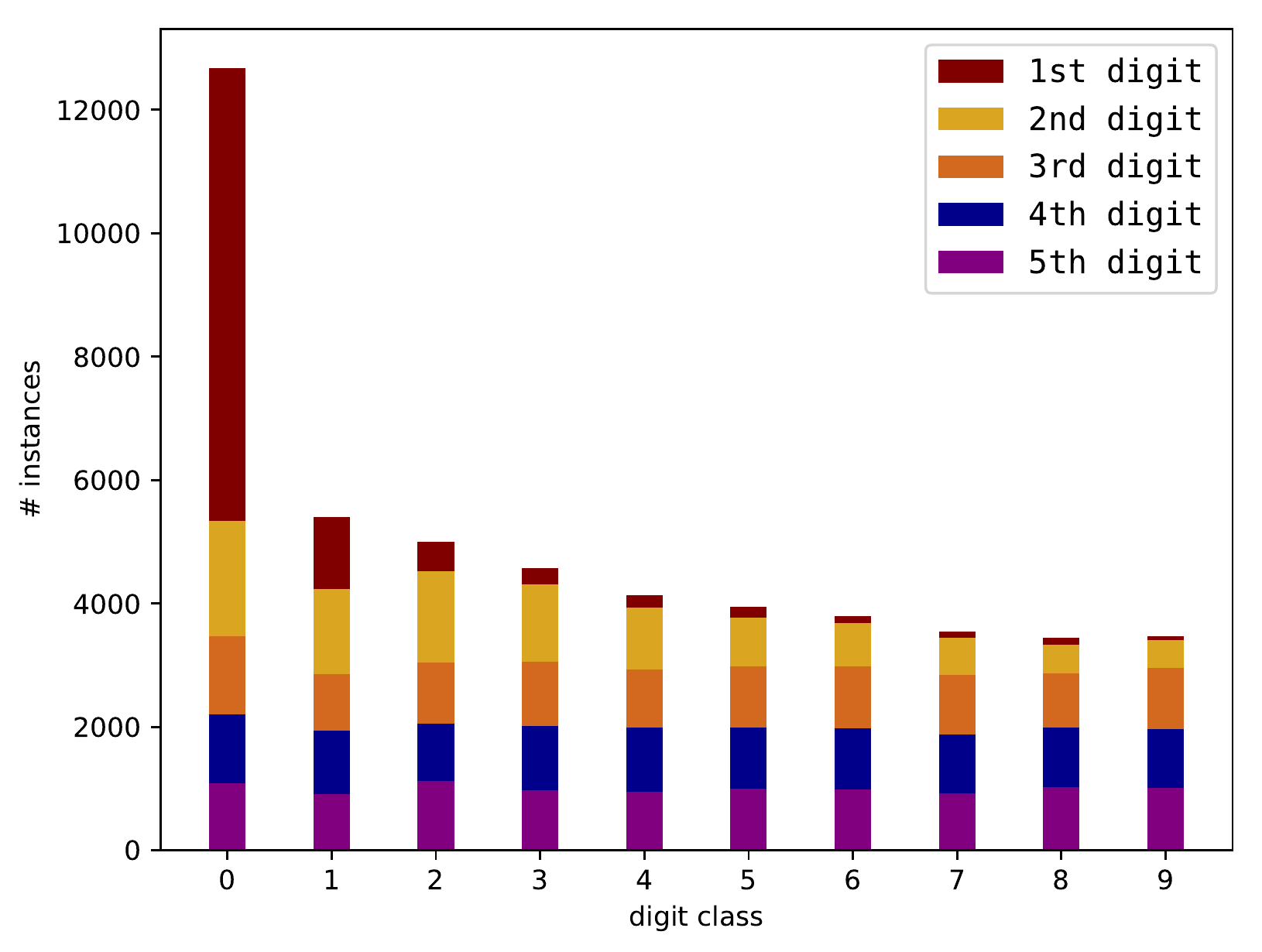}
    
    \vspace{-3mm}
    
    \caption{\small Distribution of the digit classes in the \dataset dataset. In the \gls*{amr} context, it is common that the digit `$0$' has many more instances than the others, as a brand new meter starts with \texttt{00000}.}
    \label{fig:frequency}
\end{figure}

\begin{figure*}[!htb]
    \centering
    \includegraphics[width=0.8\linewidth]{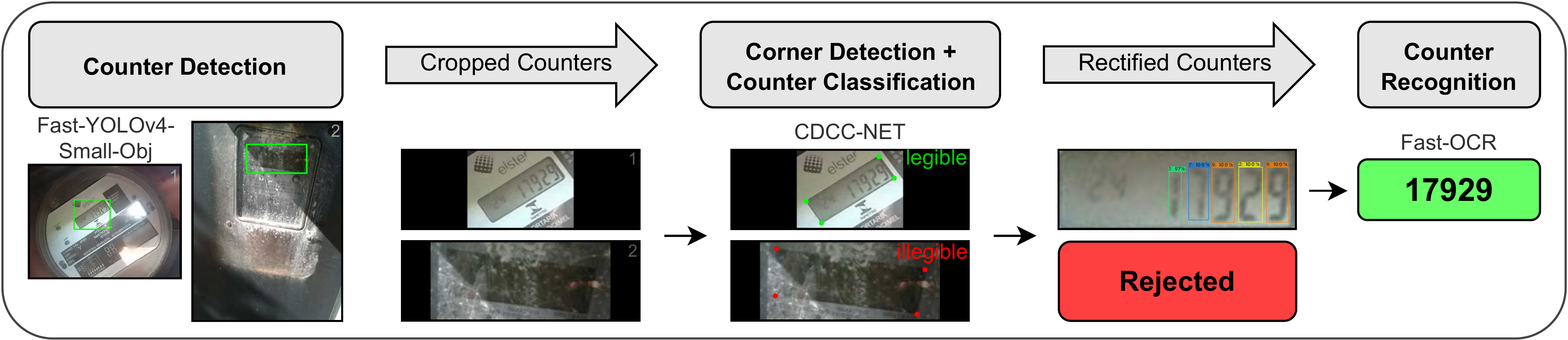}
    
    \vspace{-1.25mm}
    
    \caption{\small The pipeline of the proposed \gls*{amr} system. Given an input image, the counter region is located using a modified version of the Fast-YOLOv4 model.
    Then, in a single forward pass of the proposed CDCC-NET, the cropped counter is classified as legible/operational or illegible/faulty and the position ($x$,~$y$) of each of its corners is predicted.
    Finally, illegible counters are rejected, whereas legible ones are rectified and fed into our recognition network, called~\ocrnet.}
    \label{fig:proposed-pipeline}
    \vspace{-1mm}
\end{figure*}

In electromechanical meters, it is possible that the digits (usually, the rightmost one) are rotating (see an example in the $3$rd row and $4$th column in Fig.~\ref{fig:samples-dataset}).
In such cases, following the protocol adopted at \gls*{copel}, we considered the lowest digit as the ground truth (e.g., a digit rotating from `4' to `5' is labeled as `4'), except between digits `$9$' and `$0$' where the digit should be labeled as~`$9$'.

With the advances of deep learning-based techniques and the availability of ever larger datasets, in many cases it is time-consuming to divide the datasets multiple times and then average the results among multiple runs.
Hence, public datasets introduced in recent years commonly have a single division of the images into training, validation and test sets~\cite{laroca2019convolutional,salomon2020deep}.
In the same direction, we randomly split the \dataset dataset as follows: $5{,}000$ images for training, $5{,}000$ images for testing and $2{,}500$ images for validation, following the split protocol (i.e.,~$40$\%/$40$\%/$20$\%) used in the \ufpramr dataset.
We preserved the percentage of samples for illegible/faulty meters, that is, there are $1{,}000$ images of illegible/faulty meters in each of the training and test sets, and $500$ images in the validation~one.
For reproducibility purposes, the subsets generated are explicitly available along with the proposed~dataset.
\section{Proposed Approach}
\label{sec:proposed}

As illustrated in Fig.~\ref{fig:proposed-pipeline}, the proposed approach consists of three main stages: (i)~counter detection, (ii)~corner detection and counter classification, and (iii)~counter recognition\footnote{The entire system, i.e., the architectures and weights, is publicly available at \url{\supplementary}}.
Given an input image, the counter region is located using a modified version of the Fast-YOLOv4 model, called \detnet.
Then, in a single forward pass of the proposed \textit{\gls*{cdcc}}, the cropped counter is classified as legible/operational or illegible/faulty and the position~($x$,~$y$) of each of its corners is predicted.
Finally, illegible counters are rejected, while legible ones are rectified and fed into our recognition network, called~\ocrnet.

In the remainder of this section, each stage of the proposed system is better described.
It is worth noting that, for each stage, we train a single network on images from both datasets in which we perform experiments (see Section~\ref{sec:experiments-setup}).
In this way, our networks become robust to images captured under different conditions with significantly less manual effort, as the network parameters are adjusted only~once.

\subsection{Counter Detection}

In unconstrained scenarios, locating the corners~($2$D points) of the counter directly in the input image is a challenging task for three main reasons:
(i) one or more corners may not be visible due to occlusions caused by dirt, reflections, and other factors;
(ii)~the counter region may occupy a very small portion of the input image, as can be seen in Fig.~\ref{fig:samples-dataset}; and (iii)~some text blocks (e.g., meter specifications and serial number) are very similar to the counter region in certain meter models.
Therefore, we first locate the counter in the input image and then detect its corners in the cropped~patches.

As the counter region is rectified prior to the recognition stage in our system's pipeline, the counter detector does not need to be very sophisticated or rely on very deep models.
In other words, our \gls*{amr} system can tolerate less accurate detections of the counter region, as the corners will be later located and the counter rectified.
Taking this into account as well as the importance of having an efficient system in real-world applications, we initially  decided to use the Fast-YOLOv4 model~\cite{fastyolov4} for this task since, despite being much smaller than state-of-the-art object detectors, Fast-YOLO variants (also known as Tiny-YOLO) are still able to detect some objects quite precisely~\cite{redmon2016yolo} and have been employed in various research areas in recent years~\cite{laroca2018robust,jiang2019mixed,salomon2020deep}.

Nevertheless, we noticed in preliminary experiments that Fast-YOLOv4 failed in some cases where the meter was relatively far from the camera (usually in images where the reading is illegible).
Therefore, we performed some modifications to the network in order to improve its performance in detecting small objects.
More specifically, following insights from \cite{redmon2018yolov3,bochkovskiy2020yolov4,alexeyab}, we added a few layers to the network so that it predicts bounding boxes at $3$ different scales instead of~$2$.
This was done by
(i)~taking the feature map from the next-to-last layer and upsampling it by a factor of $2$; 
(ii)~concatenating a feature map from earlier in the network with the upsampled features; and (iii)~adding some convolutional layers to process this combined feature map and predict a similar tensor but with twice the size.
Table~\ref{tab:fast-yolov4-small-objs} shows the modified architecture, which hereinafter is referred to as \textit{\detnet}.
Observe that the final feature map is now $48\times48$ instead of $24\times24$ pixels (for an input size of $384$~$\times$~$384$~pixels), which makes fine details better visible; consequently, small objects can be detected more~accurately.
There are $18$ filters (instead of $255$) in layers $15$, $22$ and $29$ so that the network predicts $1$ class instead of~$80$.

\begin{table}[!htb]
\centering
\caption{\small The \detnet model, a modified version of Fast-YOLOv4 to improve the detection of small objects. We added layers $38$-$44$ so that the network predicts bounding boxes at $3$ different scales (layers $30$, $37$, and $44$) instead of~$2$~(layers $30$ and~$37$).}
\label{tab:fast-yolov4-small-objs}

\vspace{0.5mm}

\resizebox{0.85\columnwidth}{!}{ %
\begin{tabular}{@{}cccccc@{}}
\toprule
\textbf{\#} & \textbf{Layer} & \textbf{Filters} & \textbf{Size} & \textbf{Input} & \textbf{Output} \\ \midrule
$0$ & conv & $32$ & $3 \times 3 / 2$ & $384 \times 384 \times 3$ & $192 \times 192 \times 32$ \\
$1$ & conv & $64$ & $3 \times 3 / 2$ & $192 \times 192 \times 32$ & $96 \times 96 \times 64$ \\
$2$ & conv & $64$ & $3 \times 3 / 1$ & $96 \times 96 \times 64$ & $96 \times 96 \times 64$ \\
$3$ & route [$2$] &  &  & $\phantom{96 \times 196} \times 1/2$ & $96 \times 96 \times 32$ \\
$4$ & conv & $32$ & $3 \times 3 / 1$ & $96 \times 96 \times 32$ & $96 \times 96 \times 32$ \\
$5$ & conv & $32$ & $3 \times 3 / 1$ & $96 \times 96 \times 32$ & $96 \times 96 \times 32$ \\
$6$ & route [$5$, $4$] &  &  & & $96 \times 96 \times 64$ \\
$7$ & conv & $64$ & $1 \times 1 / 1$ & $96 \times 96 \times 64$ & $96 \times 96 \times 64$ \\
$8$ & route [$2$, $7$] &  &  & & $96 \times 96 \times 128$ \\
$9$ & max & & $2 \times 2 / 2$ & $96 \times 96 \times 128$ & $48 \times 48 \times 128$ \\
$10$ & conv & $128$ & $3 \times 3 / 1$ & $48 \times 48 \times 128$ & $48 \times 48 \times 128$ \\
$11$ & route [$10$] &  &  & $\phantom{96 \times 96} \times 1/2$ & $48 \times 48 \times 64$ \\
$12$ & conv & $64$ & $3 \times 3 / 1$ & $48 \times 48 \times 64$ & $48 \times 48 \times 64$ \\
$13$ & conv & $64$ & $3 \times 3 / 1$ & $48 \times 48 \times 64$ & $48 \times 48 \times 64$ \\
$14$ & route [$13$, $12$] &  &  & & $48 \times 48 \times 128$ \\
$15$ & conv & $128$ & $1 \times 1 / 1$ & $48 \times 48 \times 128$ & $48 \times 48 \times 128$ \\
$16$ & route [$10$, $15$] &  &  & & $48 \times 48 \times 256$ \\
$17$ & max & & $2 \times 2 / 2$ & $48 \times 48 \times 256$ & $24 \times 24 \times 256$ \\
$18$ & conv & $256$ & $3 \times 3 / 1$ & $24 \times 24 \times 256$ & $24 \times 24 \times 256$ \\
$19$ & route [$18$] &  &  & $\phantom{96 \times 96} \times 1/2$ & $24 \times 24 \times 128$ \\
$20$ & conv & $128$ & $3 \times 3 / 1$ & $24 \times 24 \times 128$ & $24 \times 24 \times 128$ \\
$21$ & conv & $128$ & $3 \times 3 / 1$ & $24 \times 24 \times 128$ & $24 \times 24 \times 128$ \\
$22$ & route [$21$, $20$] &  &  & & $24 \times 24 \times 256$ \\
$23$ & conv & $256$ & $1 \times 1 / 1$ & $24 \times 24 \times 256$ & $24 \times 24 \times 256$ \\
$24$ & route [$18$, $23$] &  &  & & $24 \times 24 \times 512$ \\
$25$ & max & & $2 \times 2 / 2$ & $24 \times 24 \times 512$ & $12 \times 12 \times 512$ \\
$26$ & conv & $512$ & $3 \times 3 / 1$ & $12 \times 12 \times 512$ & $12 \times 12 \times 512$ \\
$27$ & conv & $256$ & $1 \times 1 / 1$ & $12 \times 12 \times 512$ & $12 \times 12 \times 256$ \\
$28$ & conv & $512$ & $3 \times 3 / 1$ & $12 \times 12 \times 256$ & $12 \times 12 \times 512$ \\
$29$ & conv & $18$ & $1 \times 1 / 1$ & $12 \times 12 \times 512$ & $12 \times 12 \times 18$ \\
$30$ & \textbf{detection} &  &  &  &  \\[1.75pt] \cdashline{1-6} \\[-6pt]
$31$ & route [$27$] & & & & $12 \times 12 \times 256$ \\
$32$ & conv & $128$ & $1 \times 1 / 1$ & $12 \times 12 \times 256$ & $12 \times 12 \times 128$ \\
$33$ & upsample & & \phantom{aaaa} $2\times$ & $12 \times 12 \times 128$ & $24 \times 24 \times 128$ \\
$34$ & route [$33$, $23$] &  & &  & $24 \times 24 \times 384$ \\
$35$ & conv & $256$ & $3 \times 3 / 1$ & $24 \times 24 \times 384$ & $24 \times 24 \times 256$ \\
$36$ & conv & $18$ & $1 \times 1 / 1$ & $24 \times 24 \times 256$ & $24 \times 24 \times 18$ \\
$37$ & \textbf{detection} &  &  &  &  \\[1.75pt] \cdashline{1-6} \\[-6pt]
$38$ & route [$35$] & & & & $24 \times 24 \times 256$ \\
$39$ & conv & $64$ & $1 \times 1 / 1$ & $24 \times 24 \times 256$ & $24 \times 24 \times 64$ \\
$40$ & upsample & & \phantom{aaaa} $2\times$ & $24 \times 24 \times 64$ & $48 \times 48 \times 64$ \\
$41$ & route [$40$, $15$] &  & &  & $48 \times 48 \times 192$ \\
$42$ & conv & $128$ & $3 \times 3 / 1$ & $48 \times 48 \times 192$ & $48 \times 48 \times 128$ \\
$43$ & conv & $18$ & $1 \times 1 / 1$ & $48 \times 48 \times 128$ & $48 \times 48 \times 18$ \\
$44$ & \textbf{detection} &  &  &  &  \\ \bottomrule
\end{tabular}
} %
\end{table}

The input size of $384\times384$ pixels was chosen based on careful assessments carried out in the validation set, where we sought the best balance between speed and
accuracy with different input dimensions  (from $320\times320$ to $608\times608$ pixels).
It is remarkable that, according to our experiments, for this task, \detnet performed better than the Fast-YOLOv4 model with larger input~sizes, while requiring comparable or even less \gls*{flop} in each forward~pass.
As an example, \detnet, which requires~$6.8$ \gls*{bflop}, achieved $99.88$ \gls*{map} in the validation set, whereas Fast-YOLOv4 with an input size of $608\times608$ pixels reached~$99.13$\% \gls*{map} while requiring $14.5$ \gls*{bflop} in each forward~pass.

We exploit several data augmentation strategies to train our network, such as random cropping, shearing, conversion to grayscale, and random perturbations of hue, saturation and
brightness.
As each image contains a single meter, only the detection with the highest confidence value is considered in cases where more than one counter region is~predicted.

\subsection{\cdcc}

In order to rectify the cropped counter patch, we need to first locate the four corners of the counter.
For this purpose, we designed a \textit{multi-task} network~\cite{zhang2018survey}, called \textit{\gls*{cdcc}}, that analyzes the counter region detected in the previous stage and predicts $9$ outputs: eight float numbers referring to the corner positions ($x_0$/$w$, $y_0$/$h$, \dots, $x_3$/$w$, $y_3$/$h$) and an array containing two float numbers regarding the probability of the counter being legible/operational or illegible/faulty.
We consider as input to the network a counter region slightly larger than the one detected in the previous stage in order to try to ensure that all corners are within the cropped patch even in less accurate~detections.

\gls*{cdcc}'s architecture is shown in Table~\ref{tab:cdcc-architecture}.
As can be seen, there are three shared convolutional layers with $16$/$32$/$64$ filters, each followed by a max-pooling layer with a $2\times2$~kernel and stride~$=2$.
There are also two fully connected (or dense) layers for each of the $9$ outputs (i.e., these two layers are not shared).
Observe that in the second non-shared dense layer there is a single unit for the prediction of each of the eight corner coordinates (a single float number is predicted for each task), and two units for the prediction of the probabilities of the counter being legible or~illegible (here we employed the softmax function to enforce that the sum of the probabilities is equal~to~$1$).

\begin{table}[!htb]
\centering
\caption{\small \gls*{cdcc}'s layers and hyperparameters. It is relatively shallow and has two dense layers for each of the $9$ outputs (i.e., $x_0$/$w$, $y_0$/$h$, $x_1$/$w$, $y_1$/$h$, $x_2$/$w$, $y_2$/$h$, $x_3$/$w$, $y_3$/$h$, $[$legible, illegible$]$).}
\label{tab:cdcc-architecture}

\vspace{0.75mm}

\resizebox{0.99\columnwidth}{!}{ %
\begin{tabular}{@{}cccccc@{}}
\toprule
\textbf{\#} & \textbf{Layer} & \textbf{Filters} & \textbf{Size} & \textbf{Input} & \textbf{Output} \\ \midrule
$0$ & conv & $16$ & $3 \times 3 / 1$ & $192 \times 64 \times 3$ & $192 \times 64 \times 16$ \\
$1$ & max &  & $2 \times 2 / 2$ & $192 \times 64 \times 16$ & $96 \times 32 \times 16$ \\
$2$ & conv & $32$ & $3 \times 3 / 1$ & $96 \times 32 \times 16$ & $96 \times 32 \times 32$ \\
$3$ & max &  & $2 \times 2 / 2$ & $96 \times 32 \times 32$ & $48 \times 16 \times 32$ \\
$4$ & conv & $64$ & $3 \times 3 / 1$ & $48 \times 16 \times 32$ & $48 \times 16 \times 64$ \\
$5$ & max & & $2 \times 2 / 2$ & $48 \times 16 \times 64$ & $24 \times 8 \times 64$ \\
$6$ & flatten & & & $24 \times 8 \times 64$ & $12288$ \\

\toprule
\textbf{\#} & \textbf{Layer} & \textbf{Connected to} & \textbf{Units} & \textbf{Input} & \textbf{Output} \\ \midrule

$7$ & non-shared dense$_{[0..8]}$ & $\# 6$ & $128$ & $12288$ & $128$ \\[2.75pt] \cdashline{1-6} \\[-7pt]
$8$ & non-shared dense\_corners$_{[0..7]}$ & $\# 7$ & $1$ & $128$ & $1$ \\
$9$ & non-shared dense\_counter\_class$_{[8]}$ & $\# 7$ & $2$ & $128$ & $2$ \\ \bottomrule

\end{tabular}
} %
\end{table}

The input size of the \gls*{cdcc} model is $192\times64$ pixels.
These dimensions were defined by halving the input size used in the previous stage, as here the region of interest is already cropped, and by adapting it to the mean aspect ratio of the counters in the \dataset dataset~($w$/$h\approx3$).
Thus, all images are resized to $192\times64$ pixels before being fed into the network. 
However, to avoid distortions when resizing the images, we first add black borders on them so that they have an aspect ratio~($w$/$h$) close/equal to~$3$.

The main difference between \gls*{cdcc} and existing networks for corner detection in other applications, such as license plate recognition~\cite{meng2018robust,yoo2020deep}, is that the proposed network is relatively shallow and has a specific dense layer for predicting each output, while the existing models usually have more intermediate layers with many more filters and a single dense layer to predict all output~values.
\major{Another approach worth mentioning is that proposed by Lyu et al.}~\cite{lyu2018multi} \major{for multi-oriented scene text detection, where each corner point is redefined and represented by a horizontal square $C$ = ($x_c$, $y_c$, $ss$, $ss$), where $x_c$, $y_c$ are the coordinate of a corner point (such as $x_1$, $y_1$ for top-left point) as well as the center of the horizontal square ($ss$ is the length short side of the rotated rectangular bounding box).
Then, the corner points are detected as default bounding boxes through a model with a backbone adapted from VGG16}~\cite{simonyan2015vgg} \major{containing several extra convolutional layers and a few deconvolution~modules.}

Considering that the number of training images is still limited to train such a multi-task network (i.e., there is no public dataset for \gls*{amr} with hundreds of thousands of labeled images) and also the fact that the counter region is well-aligned in most cases (especially in the \ufpramr dataset~\cite{laroca2019convolutional}), we created many artificial images through data augmentation in order to prevent overfitting.
We performed random variations of hue, saturation and brightness to the original images, in addition to randomly rotating and cropping~them.
We also randomly permuted the position of the digits on the counters to eliminate undesirable biases in network learning related to the corner positions and certain classes of digits, for example, the network might learn a false correlation between the top-left/bottom-left corners and digits `$0$' since most occurrences of the class `$0$' are in the leftmost digit position (see Fig.~\ref{fig:frequency}).
Fig.~\ref{fig:data-aug-preprocessing-ocr} shows some examples of the images generated by~us.
\major{Note that the bounding box of each digit, which is labeled in our dataset, is required to apply this data augmentation~technique.}

\begin{figure}[!htb]
    \centering
    
    \resizebox{0.99\linewidth}{!}{ %
    \includegraphics[width=0.24\linewidth]{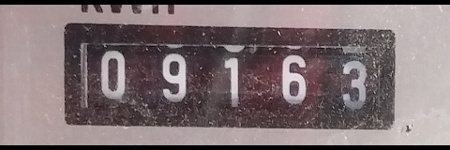}
    \includegraphics[width=0.24\linewidth]{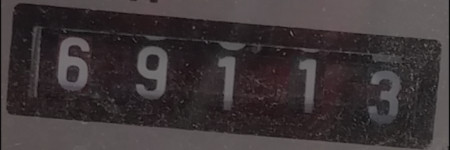}
    \includegraphics[width=0.24\linewidth]{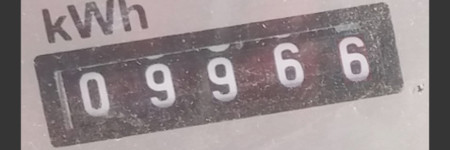}
    \includegraphics[width=0.24\linewidth]{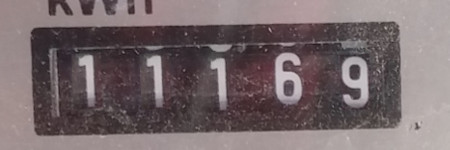}
    } %
    
    \vspace{0.35mm}
    
    \resizebox{0.99\linewidth}{!}{ %
    \includegraphics[width=0.24\linewidth]{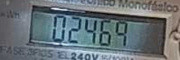}
    \includegraphics[width=0.24\linewidth]{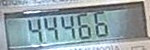}
    \includegraphics[width=0.24\linewidth]{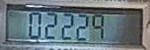}
    \includegraphics[width=0.24\linewidth]{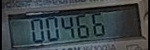}
    } %
    
    \vspace{0.35mm}
    
    \resizebox{0.99\linewidth}{!}{ %
    \includegraphics[width=0.24\linewidth]{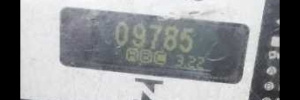}
    \includegraphics[width=0.24\linewidth]{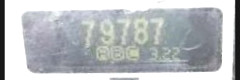}
    \includegraphics[width=0.24\linewidth]{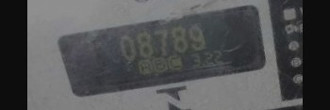}
    \includegraphics[width=0.24\linewidth]{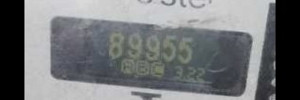}
    } %
    
    \vspace{0.35mm}
    
    \resizebox{0.99\linewidth}{!}{ %
    \includegraphics[width=0.24\linewidth]{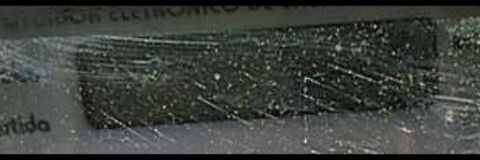}
    \includegraphics[width=0.24\linewidth]{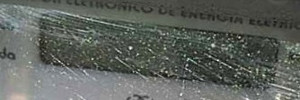}
    \includegraphics[width=0.24\linewidth]{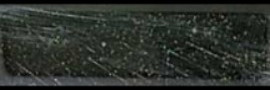}
    \includegraphics[width=0.24\linewidth]{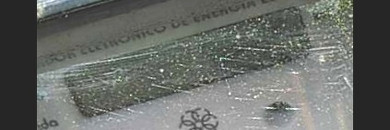}
    } %
    
    \vspace{-1.5mm}
    
    \caption{\small Some representative examples of the images generated by us for training \gls*{cdcc} \major{(and also Fast-OCR; see the next section).}
    The images in the first column are the originals and the others were generated~automatically.}
    \label{fig:data-aug-preprocessing-ocr}
\end{figure}


\major{After locating the four corners, we rectify the counters classified as legible/operational by calculating and applying a perspective transform from the coordinates of the four corners in the source image~(src) to the corresponding vertices in the ``unwarped'' image~(dst).
We defined these vertices as follows: ($0$, $0$)  indicates the top-left corner; ($max_w-1$, $0$)  corresponds to the top-right corner; ($max_w-1$, $max_h-1$)  is the bottom-right corner; and ($0$, $max_h-1$) refers to the bottom-left corner, where $max_w$ indicates the maximum distance between the bottom-right and bottom-left $x$ coordinates or the top-right and top-left $x$ coordinates, and $max_h$ is the maximum distance between the top-right and bottom-right $y$ coordinates or the top-left and bottom-left $y$~coordinates.}

\major{For the sake of completeness, following~}\cite{opencv_transformations}\major{, the $3\times3$ matrix of the perspective transform is calculated so that:}

\begin{equation}
\begin{bmatrix} 
    t_ix'_i \\ t_iy'_i \\ t_i
\end{bmatrix} = \text{map\_matrix} \cdot \begin{bmatrix} 
    x_i \\ y_i \\ 1 
\end{bmatrix}
\end{equation}

\noindent \major{where}
\vspace{-1mm}
\begin{equation}
    dst(i)=(x'_i,y'_i), src(i)=(x_i,y_i),  i=0,1,2,3 
\end{equation}

\noindent \major{and the counter region is rectified using the specified matrix:}
\vspace{-1mm}
\begin{equation}
    dst(x,y)=src \begin{pmatrix}
\frac{M_{11}x + M_{12}y + M_{13}}{M_{31}x + M_{32}y + M_{33}} , \frac{M_{21}x + M_{22}y + M_{23}}{M_{31}x + M_{32}y + M_{33}}
\end{pmatrix}
\end{equation}


Counters classified by \gls*{cdcc} as illegible/faulty, on the other hand, are rejected.
To the best of our knowledge, in the \gls*{amr} context, this is the first work in which the region of interest is rectified prior to the recognition stage.
As illustrated in Fig.~\ref{fig:counter-rectification-before-after}, rectified counters become more horizontal, tightly-bounded, and easier to~read.

\begin{figure}[!htb]
    \centering
    \captionsetup[subfigure]{captionskip=2pt,font={scriptsize},justification=centering} 
    
	\subfloat[][detected counter regions]{
    \includegraphics[width=0.35\linewidth]{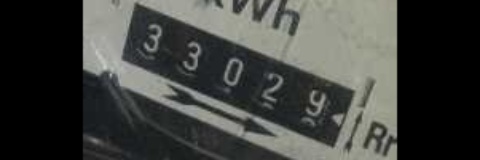}
    \includegraphics[width=0.35\linewidth]{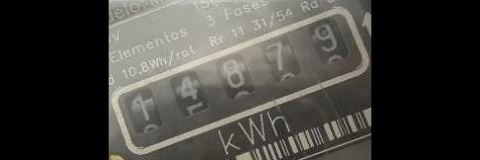}} 
    
    \vspace{1.1mm}
    
	\subfloat[][rectified counter regions]{
    \includegraphics[width=0.35\linewidth]{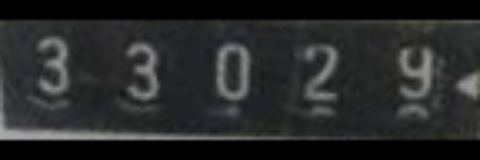}
    \includegraphics[width=0.35\linewidth]{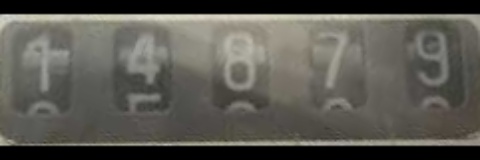}} 
    
    \vspace{-1.25mm}
    
    \caption{\small Two examples of counter regions before and after the rectification process. It should be observed that the rectified counters become more horizontal, tightly-bounded, and easier to~read.}
    \label{fig:counter-rectification-before-after}
\end{figure}

\subsection{Counter Recognition}

Once the counter region has been located and rectified, the digits must be recognized.
To this end, we built a new \major{lightweight} detection network, called \textit{\ocrnet}, that incorporates features from existing models focused on the speed/accuracy trade-off, such as YOLOv2~\cite{redmon2017yolo9000}, CR-NET~\cite{silva2017realtime} and Fast-YOLOv4.
In this work, we handle counter recognition as an object detection problem since object detectors have been successfully applied to several character recognition tasks in recent years~\cite{hochuli2020endtoend,silva2020realtime,laroca2021efficient}.
Accordingly, the \ocrnet model is trained to predict $10$ classes (i.e., $0$-$9$) using the cropped counter patch as well as the class and bounding box~($x$, $y$, $w$, $h$) of each digit as inputs.

The architecture of \ocrnet is shown in Table~\ref{tab:fast-ocr}.
The input size is $384\times128$~pixels considering both the number of max-pooling layers in the network (i.e., the dimensions of the output layer must be large enough to enable the detection of multiple digits horizontally spread side by side) and also the mean aspect ratio of the counters in the \dataset dataset~($w$/$h\approx3$).
As Fast-YOLOv4, the proposed network performs detection at $2$~different scales (layers 14 and 21).
This is particularly important for counter recognition in unconstrained scenarios due to the fact that the digits may occupy either a small or a large portion of the counter region depending on the meter model, as illustrated in Fig.~\ref{fig:ocr-two-scales}, and also on how accurately the counter corners were detected in the previous stage.
As in the networks introduced in~\cite{redmon2017yolo9000,silva2017realtime}, the convolutional layers in \ocrnet mostly have $3\times3$~kernels and the number of filters is doubled after each max-pooling layer.
In addition, there are $1\times1$~convolutional layers between $3\times3$ convolutions to reduce the feature space from preceding~layers.

\begin{table}[!htb]
\centering
\caption{\small The architecture of the \ocrnet model, proposed for counter recognition.}
\label{tab:fast-ocr}

\vspace{0.5mm}

\resizebox{0.95\columnwidth}{!}{ %
\begin{tabular}{@{}ccccccc@{}}
\toprule
\textbf{\#} & \textbf{Layer} & \textbf{Filters} & \textbf{Size} & \textbf{Input} & \textbf{Output} & \textbf{BFLOP} \\ \midrule
$0$ & conv & $32$ & $3 \times 3 / 1$ & $384 \times 128 \times 3$ & $384 \times 128 \times 32$ & $0.085$ \\
$1$ & max &  & $2 \times 2 / 2$ & $384 \times 128 \times 32$ & $192 \times 64 \times 32$ & $0.002$ \\
$2$ & conv & $64$ & $3 \times 3 / 1$ & $192 \times 64 \times 32$ & $192 \times 64 \times 64$ & $0.453$ \\
$3$ & max &  & $2 \times 2 / 2$ & $192 \times 64 \times 64$ & $96 \times 32 \times 64$ & $0.001$ \\
$4$ & conv & $128$ & $3 \times 3 / 1$ & $96 \times 32 \times 64$ & $96 \times 32 \times 128$ & $0.453$ \\
$5$ & max & & $2 \times 2 / 2$ & $96 \times 32 \times 128$ & $48 \times 16 \times 128$ & $0.000$ \\
$6$ & conv & $256$ & $3 \times 3 / 1$ & $48 \times 16 \times 128$ & $48 \times 16 \times 256$ & $0.453$ \\
$7$ & conv & $128$ & $1 \times 1 / 1$ & $48 \times 16 \times 256$ & $48 \times 16 \times 128$ & $0.050$ \\
$8$ & conv & $256$ & $3 \times 3 / 1$ & $48 \times 16 \times 128$ & $48 \times 16 \times 256$ & $0.453$ \\
$9$ & max & & $2 \times 2 / 2$ & $48 \times 16 \times 256$ & $24 \times 8 \times 256$ & $0.000$ \\
$10$ & conv & $512$ & $3 \times 3 / 1$ & $24 \times 8 \times 256$ & $24 \times 8 \times 512$ & $0.453$ \\
$11$ & conv & $256$ & $1 \times 1 / 1$ & $24 \times 8 \times 512$ & $24 \times 8 \times 256$ & $0.050$ \\
$12$ & conv & $512$ & $3 \times 3 / 1$ & $24 \times 8 \times 256$ & $24 \times 8 \times 512$ & $0.453$ \\
$13$ & conv & $45$ & $1 \times 1 / 1$ & $24 \times 8 \times 512$ & $24 \times 8 \times 45$ & $0.009$ \\
$14$ & \textbf{detection} &  &  & & & \\
$15$ & route [$11$] &  &  & & $24 \times 8 \times 256$ & \\
$16$ & conv & $256$ & $1 \times 1 / 1$ & $24 \times 8 \times 256$ & $24 \times 8 \times 256$ & $0.025$ \\
$17$ & upsample & & \phantom{aaaa} $2\times$ & $24 \times 8 \times 256$ & $48 \times 16 \times 256$ & \\
$18$ & route [$17$, $6$] &  & &  & $48 \times 16 \times 512$ & \\
$19$ & conv & $512$ & $3 \times 3 / 1$ & $48 \times 16 \times 512$ & $48 \times 16 \times 512$ & $3.624$ \\
$20$ & conv & $45$ & $1 \times 1 / 1$ & $48 \times 16 \times 512$ & $48 \times 16 \times 45$ & $0.035$ \\
$21$ & \textbf{detection} &  &  &  & \\ \bottomrule
\end{tabular}
} %
\end{table}

\begin{figure}[!htb]
    \centering
    \captionsetup[subfigure]{captionskip=2pt,font={scriptsize},justification=centering} 
    
    \subfloat[][digits occupying a large portion of the counter]{
    \includegraphics[width=0.35\linewidth]{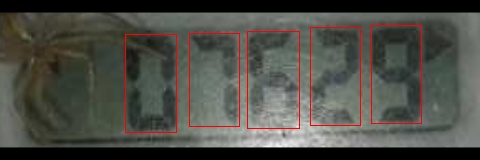}
    \includegraphics[width=0.35\linewidth]{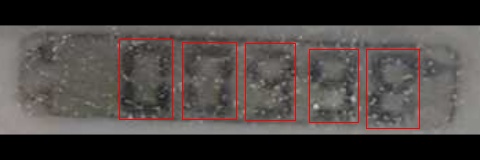}}
    
    \vspace{1.1mm}
    
    \subfloat[][digits occupying a small portion of the counter]{
    \includegraphics[width=0.35\linewidth]{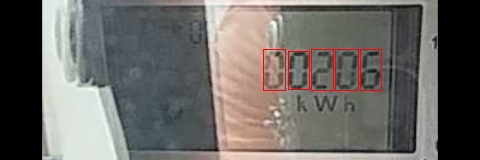}
    \includegraphics[width=0.35\linewidth]{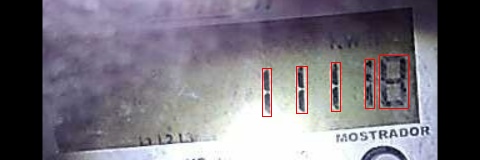}
    }
    
    \vspace{-1.25mm}
    
    \caption{\small The digits may occupy either a large~(a) or a small~(b) portion of the counter region depending on the meter model. The bounding boxes of the digits are outlined in red for better~viewing.}
    \label{fig:ocr-two-scales}
\end{figure}

In Table~\ref{tab:fast-ocr}, we also list the number of \gls*{flop} required in each layer to highlight how small the \ocrnet network is compared to some deeper detection models.
For example, for this task, our network requires $6.6$ \gls*{bflop} while YOLOv3~\cite{redmon2018yolov3} and YOLOv4~\cite{bochkovskiy2020yolov4} require $65.4$ and $59.5$ \gls*{bflop}, respectively.

It is important to note that, thanks to the versatility and ability of detection networks to learn the general features of objects (here, the digits) regardless of their positions, \major{and also to the confidence value tied to each prediction}, \ocrnet's output can be easily adapted/improved (i.e., without making any changes to its architecture) through post-processing heuristics.
For example, a variable number of digits is predicted for each counter patch fed into the network, and some digits can be discarded based on the confidence values which they were predicted or through geometric constraints; in other words, \ocrnet can be applied --~without any modification~-- to counters with different numbers of digits.
In fact, \major{we believe that} heuristic rules can also be explored to identify illegible/faulty meters erroneously classified as legible/operational in the previous~stage, \major{since it is very likely that counters classified as legible/operational should have been classified as illegible/faulty in cases where Fast-OCR has not predicted any digit with a high confidence~value.}

As the \ocrnet model is trained from scratch, many training samples are needed for the network to generalize well.
Therefore, in addition to using the original images of the counter region, we exploit the images artificially generated in the previous stage to train \ocrnet and improve its robustness.
Note that using the exactly same images in both stages is possible since patches of the counter region are fed as input into both \gls*{cdcc} and \ocrnet~models.
\section{Experiments}
\label{sec:experiments}

\subsection{Setup and Baselines}
\label{sec:experiments-setup}

In this work, we conducted experiments on images from the \dataset and \ufpramr~\cite{laroca2019convolutional} datasets.
The \ufpramr dataset contains $2{,}000$ images acquired in relatively well-controlled environments, with a resolution between $2{,}340$~$\times$~$4{,}160$ and $3{,}120$~$\times$~$4{,}160$ pixels, and is split into three subsets: training ($800$ images), validation ($400$ images) and testing ($800$ images).
In order to train, validate and test our networks, we merge the respective subsets from both datasets (i.e., exactly the same networks are used regardless of which datasets we are running experiments on).
As the \ufpramr dataset does not have any annotations related to the corners of the counters, we manually labeled their positions in its $2{,}000$ images so that we can use images from both datasets to train/evaluate the \gls*{cdcc} model.
All annotations made by us are publicly available to the research community along with the \dataset~dataset. 

It is important to point out that in several recent works in the literature~\cite{gomez2018cutting,koscevic2018automatic,calefati2019reading,liao2019reading} the authors reported only the results obtained by the proposed methods or compared them exclusively with traditional approaches, overlooking deep learning-based approaches designed for \gls*{amr}.
This makes it difficult to make a fair comparison between recently published~works.
Taking this into account, in this work, the end-to-end results achieved by the proposed system are compared (both in terms of recognition rate and execution time) with those obtained by several baseline methods~\cite{gomez2018cutting,calefati2019reading,liao2019reading,laroca2019convolutional,bochkovskiy2020yolov4} trained by us on \textit{exactly} the same images as the proposed method\footnote{The architectures and weights of the baselines implemented/trained by us are also publicly available at \url{\supplementary}}.
These specific methods, described in Section~\ref{sec:related_work}, were chosen/implemented by us for two main reasons: 
(i)~they were recently employed in the context of \gls*{amr} (except~\cite{bochkovskiy2020yolov4}, which was recently introduced) with promising/impressive results being reported, and (ii)~we believe we have the knowledge necessary to train/adjust them in the best possible way in order to ensure fairness in our experiments, as the authors provided enough details about the  architectures used, and also because we designed/employed similar networks (even the same ones in certain cases) in recent works in the context of license plate recognition and related areas~\cite{goncalves2018realtime,goncalves2019multitask,laroca2021efficient}.
Note that, in our experiments, we adapted all networks so that their input layers have the same aspect ratio~($w$/$h$~=~$3$).

Regarding the baselines, Gómez et al.\cite{gomez2018cutting} evaluated their recognition network on a dataset containing mostly images where the counter is well centered and occupies a good portion of the image; therefore, in our experiments we first detect the counter region in the input image with the \detnet model and then apply their network to the detected region.
The same was done with the recognition network proposed by Calefati et al.~\cite{calefati2019reading}, as they dealt with the counter detection stage using an \gls*{fcn} for semantic segmentation~\cite{long2015fully} that we do not have the knowledge necessary to train/adjust in the best possible way.
We consider reimplementing/retraining~it out of scope for this work since our detection model (i.e.,~\detnet) is able to achieve high F-measure rates in both datasets (see Section~\ref{sec:results:cd}) and also due to the fact that the segmentation task is generally more time-consuming than the detection~one.

The YOLO-based models were trained using the Darknet framework\footnote{\url{https://github.com/AlexeyAB/darknet/}}, while the other models were trained using Keras\footnote{\url{https://keras.io/}}. 
In Darknet, the following parameters were used:~$65K$ iterations (max batches), batch size~=~$64$, and learning rate~=~[$10$\textsuperscript{-$3$},~$10$\textsuperscript{-$4$},~$10$\textsuperscript{-$5$}] with decay steps at $26K$ and $45.5K$~iterations.
In Keras, we employed the following parameters: initial learning rate~=~$10$\textsuperscript{-$3$} (with \textit{ReduceLROnPlateau}'s patience = 3 and factor = $10$\textsuperscript{-$1$}), batch size~=~$128$, max epochs~=~100, and patience~=~$7$ (patience refers to the number of epochs with no improvement after which training will be stopped).
All networks were trained using the \gls*{sgd} optimizer, and all experiments were carried out on a computer with an AMD Ryzen Threadripper $1920$X $3.5$GHz CPU, $48$~GB of RAM, SSD~(read: $535$~MB/s; write: $445$~MB/s), and an NVIDIA Titan~V~GPU. 
We remark that all parameter values were defined based on experiments performed in the validation~set.

\major{In addition to the baselines mentioned above, we trained and evaluated all models for scene text recognition implemented in}~\cite{clovaai} (\major{i.e., CRNN}~\cite{shi2017endtoend}, \major{RARE}~\cite{shi2016robust}, \major{R2AM}~\cite{lee2016recursive}, \major{STAR-Net}~\cite{liu2016starnet}, \major{GRCNN}~\cite{wang2017deep}, \major{Rosetta}~\cite{borisyuk2018rosetta} \major{and TRBA}~\cite{baek2019what})\major{, which is the open source repository (PyTorch\footnote{\url{https://pytorch.org/}}) of Clova AI Research used to record the 1st place of ICDAR2013 focused scene text and ICDAR2019~ArT, and 3rd place of ICDAR2017COCO-Text and ICDAR2019 ReCTS~(task1).
For reasons of space and clarity, instead of reporting the results obtained by all these models, we included in our overall evaluation only the results obtained by the model that performed faster (CRNN}~\cite{shi2017endtoend}\major{) and by the one that obtained the best recognition rates~(TRBA}~\cite{baek2019what}\major{) in our~experiments.}

It is worth noting that, as stated in Section~\ref{sec:related_work}, recognition models trained exclusively on images from datasets for general robust reading (e.g., ICDAR 2013~\cite{karatzas2013icdar}) are likely to fail in AMR scenarios due to some domain-specific characteristics (e.g., rotating digits).
However, we believe that first pre-training them using images from large-scale scene text recognition datasets and then fine-tuning them on images from AMR datasets can enable  even better results to be~achieved.

\subsection{Results and Discussion}

In this section, we report the experiments carried out to verify the effectiveness of the proposed \gls*{amr} system.
We first assess the counter detection stage separately since the regions used in the following stages are extracted from the detection results, rather than cropped directly from the ground truth --~note that a detection failure probably leads to another failure in the subsequent stages.
Similarly, we then report the results reached by \gls*{cdcc} in both corner detection and counter classification tasks.
Finally, we evaluate the entire \gls*{amr} system in an end-to-end manner \major{(without any prior knowledge as to which dataset each test image belongs to or whether it is from a legible/operational or illegible/faulty meter)} and compare the reading results achieved in legible/operational images with those obtained by \numbaselines baseline methods.



\subsubsection{Counter Detection}
\label{sec:results:cd}

Detection tasks in the \gls*{amr} context are often evaluated by considering a predicted bounding box to be correct if its \gls*{iou} with the ground truth is greater than $50$\%~\cite{laroca2019convolutional,salomon2020deep}.
Nevertheless, such a low threshold~(\gls*{iou}~$> 0.5$) was deliberately defined by Everingham et al.~\cite{everingham2010pascalvoc} to account for inaccuracies in bounding boxes in the training data, as defining the bounding box for a highly non-convex object (e.g., a person with arms and legs spread) is somewhat subjective.
Taking into account that the counters are convex objects and that they were carefully labeled in both datasets, in Table~\ref{tab:results-counter-detection} we report the performance (in terms of F-measure) of the \detnet model over different \gls*{iou} thresholds, from $0.5$ to~$0.95$, similarly to the COCO~\cite{lin2014microsoft} primary metric (\gls*{map}@\gls*{iou}=[$0.5$:$0.95$]).
As we consider only one meter per image, the precision and recall rates are~identical.

\begin{table}[!htb]
\centering
\setlength{\tabcolsep}{5pt}
\caption{
\small F-measure values obtained over different \gls*{iou} thresholds, from $0.5$ to $0.95$, in the counter detection stage. 
Note that \detnet achieves considerably better results at higher \gls*{iou} thresholds (i.e., $0.8$-$0.95$), which indicates that its predictions are much better aligned with the ground~truth.
}
\label{tab:results-counter-detection}

\vspace{0.5mm}

\resizebox{0.99\linewidth}{!}{ %
\begin{tabular}{@{}lccccccc@{}}
\toprule
\multicolumn{1}{c}{\multirow{2}{*}{Model}}         & \multicolumn{7}{c}{\gls*{iou} Threshold}            \\                          & $0.5$     & $0.6$     & $0.7$    & $0.8$    & $0.9$ & $0.95$ & $0.5$:$0.95$    \\ \midrule
\small \textit{\textbf{\ufpramr~\cite{laroca2019convolutional}}} \normalsize &         &         &        &        & & &        \\
\multicolumn{1}{c}{\major{Fast-YOLOv4 ($608\times608$)}}                  & \major{$98.1$\%}  & \major{$97.8$\%}  & \major{$97.6$\%} & \major{$94.6$\%} & \major{$75.9$\%} & \major{$38.6$\%} & \major{$83.8$\%} \\
\multicolumn{1}{c}{\detnet}               & \major{$99.9$\%} & \major{$99.9$\%} & \major{$99.9$\%} & \major{$98.1$\%} & \major{$80.0$\%} & \major{$38.9$\%} & \major{$86.1$\%} \\ \midrule
\small \textit{\textbf{\dataset}} \normalsize &         &         &        &        &  & &       \\
\multicolumn{1}{c}{\major{Fast-YOLOv4 ($608\times608$)}}                    & \major{$99.0$\%}  & \major{$98.3$\%}  & \major{$95.3$\%} & \major{$84.2$\%} & \major{$52.2$\%} & \major{$17.0$\%} & \major{$74.3$\%} \\
\multicolumn{1}{c}{\detnet}             & \major{$99.7$\%}  & \major{$99.5$\%}  & \major{$98.8$\%} & \major{$96.1$\%} & \major{$75.3$\%} & \major{$28.8$\%} & \major{$83.0$\%} \\ \bottomrule
\end{tabular}
} %
\end{table}

For comparison purposes, we also list in Table~\ref{tab:results-counter-detection} the detection results obtained by the original Fast-YOLOv4 model. 
Observe that the results achieved by \detnet are considerably better at higher \gls*{iou} thresholds (i.e., $0.8$-$0.95$), which indicates that the bounding boxes predicted by the modified architecture are much better aligned with the ground~truth.
This is relevant since although our \gls*{amr} system can tolerate less accurate detections at this stage, such imprecise predictions may still impair counter rectification and, consequently, counter recognition because one or more corners may not be within the detected bounding box.
Considering the detections with \gls*{iou}~$> 0.5$ with the ground truth as correct, as in some previous works, the \detnet \major{ failed in only one image from  the UFPR-AMR's test set and in only $14$ of the $5{,}000$ test images of the Copel-AMR dataset.}
Nevertheless, we highlight that it is still possible to correctly perform the subsequent tasks in most cases where our network has failed at this stage, as the four corners are usually within the detected region (especially considering that we use as input to the next stage a counter region slightly larger than the one~detected).
\major{For the record, we tried to further improve the results achieved at this stage by adding attention modules to each model; however, Fast-YOLOv4's results improved only marginally whereas Fast-YOLOv4-SmallObj's results have not improved at~all.}

Some detection results are shown in Fig.~\ref{fig:results-counter-detection}.
As can be seen, well-located predictions were attained on meters of different models and on images acquired under unconstrained conditions, that is, with significant reflections, rotations, and scale~variations.

\begin{figure}[!htb]
    \centering
    
    \resizebox{0.995\linewidth}{!}{ %
    \includegraphics[width=0.16\linewidth]{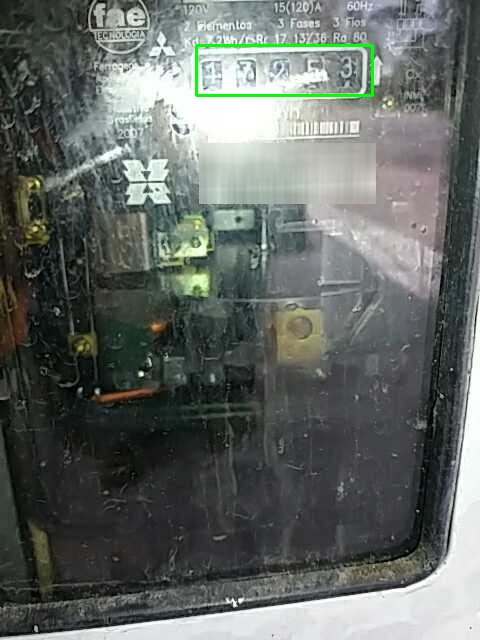}
    \includegraphics[width=0.16\linewidth]{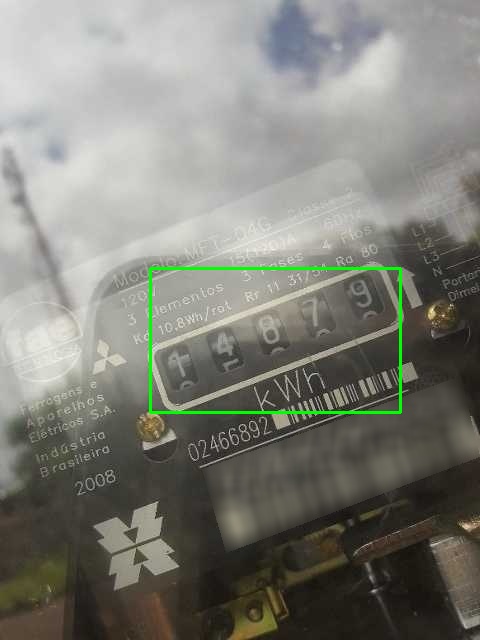}
    \includegraphics[width=0.16\linewidth]{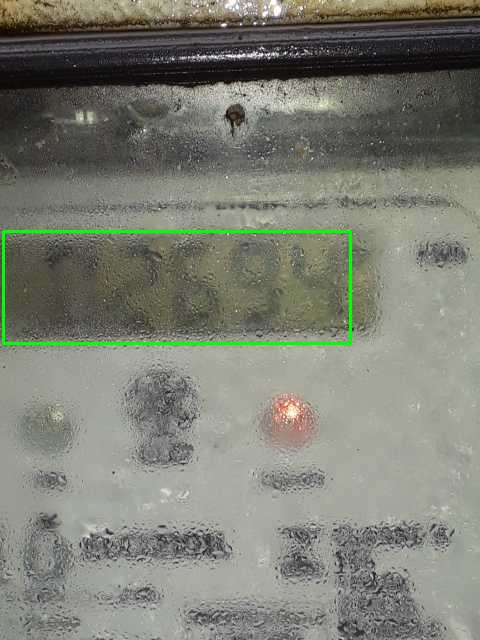}
    \includegraphics[width=0.16\linewidth]{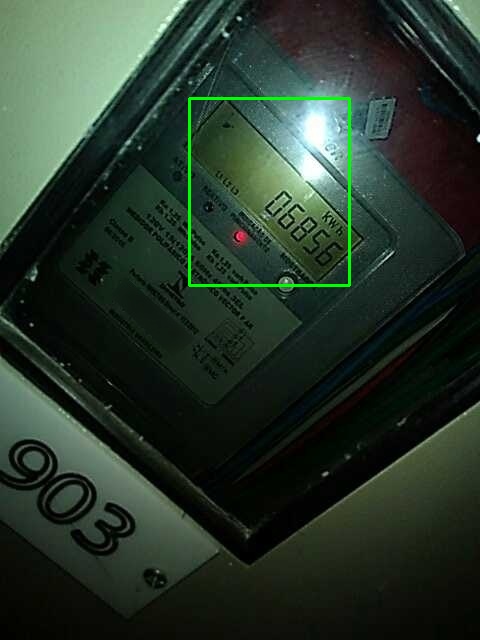}
    \includegraphics[width=0.16\linewidth]{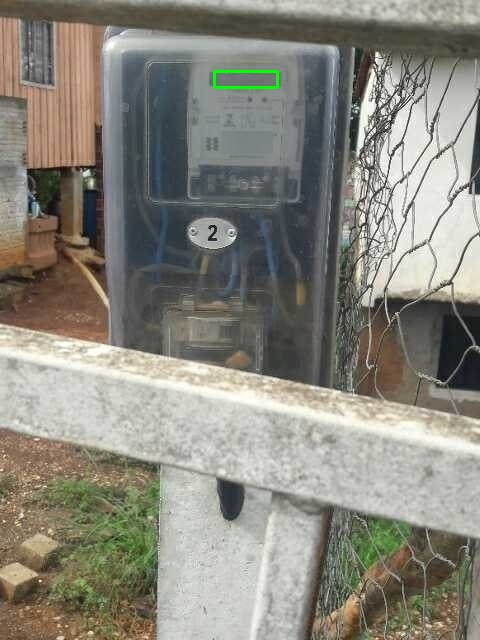}
    \includegraphics[width=0.16\linewidth]{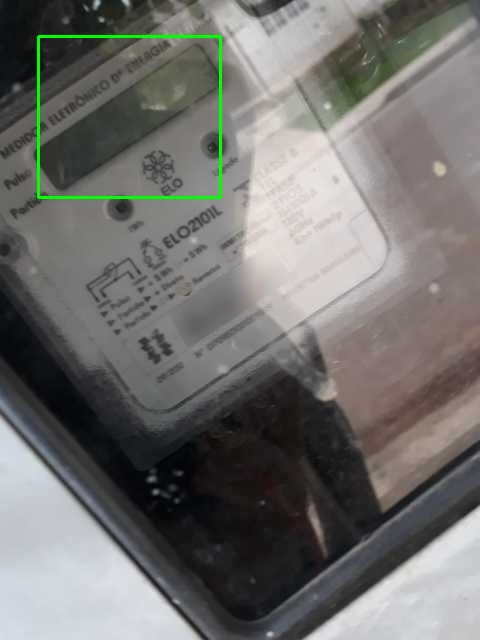}
    } %
    
    \vspace{0.5mm}
    
    \resizebox{0.995\linewidth}{!}{ %
    \includegraphics[width=0.19\linewidth]{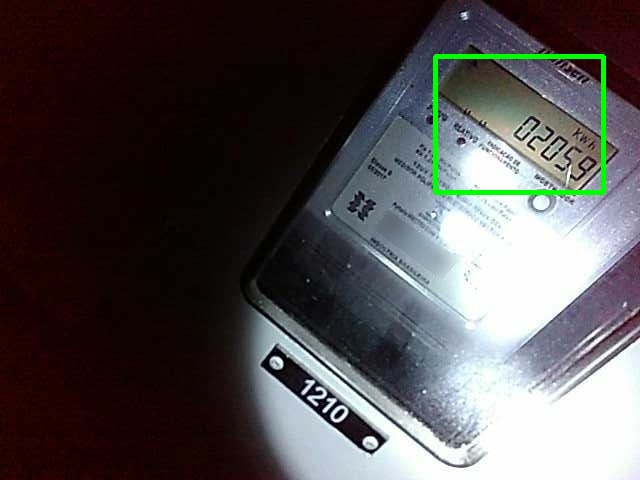}
    \includegraphics[width=0.19\linewidth]{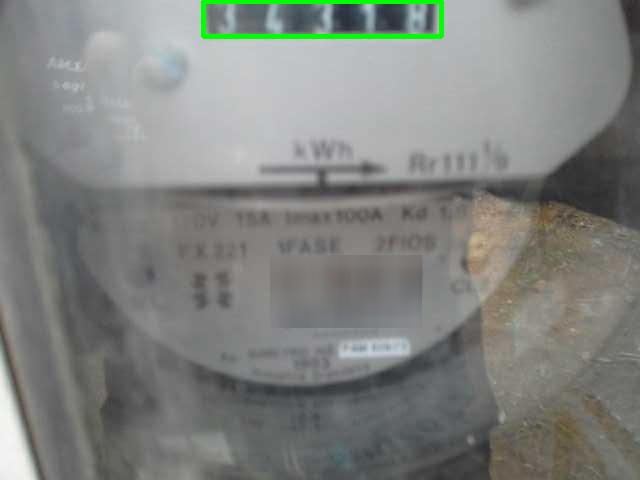}
    \includegraphics[width=0.19\linewidth]{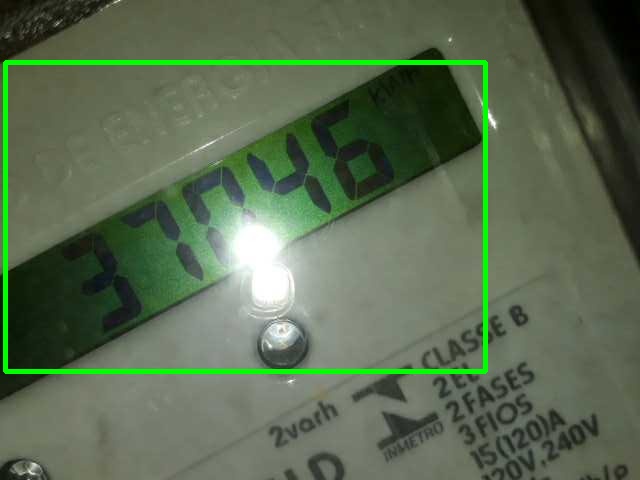}
    \includegraphics[width=0.19\linewidth]{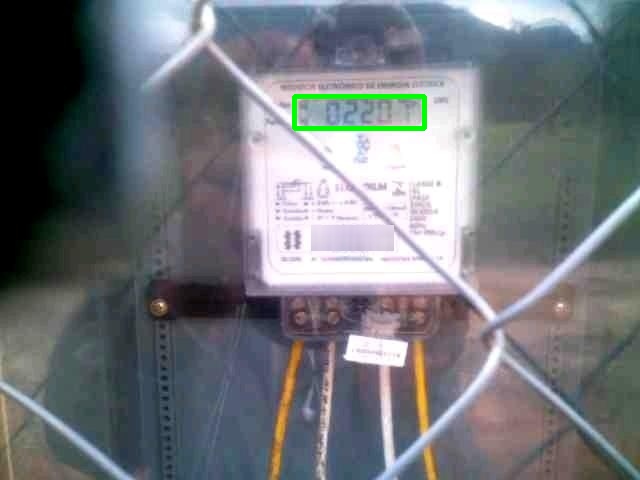}
    \includegraphics[width=0.19\linewidth]{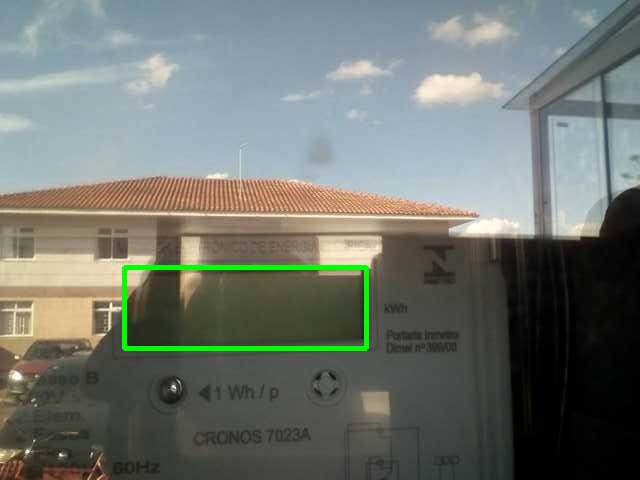}
    } %
    
    \vspace{-1mm}
    
    \caption{\small Representative samples of counters detected by the \detnet model -- images taken under unconstrained conditions, with significant reflections, rotations, and scale variations.}
    \label{fig:results-counter-detection}
\end{figure}

\subsubsection{Corner Detection and Counter Classification}
\label{sec:results:corner-detection}

To assess the performance of \gls*{cdcc} in the \textit{corner detection} task, following~\cite{yoo2020deep}, we report in Table~\ref{tab:results-corner-detection} the mean pixel distance between the predicted corner positions and the ground truth on each dataset, in addition to how many \gls*{fps} the proposed model is capable of processing (we report the average across $\nruns$ runs).
We normalize the distances by dividing them by the respective image dimensions.
To enable a comparative evaluation, we also list in Table~\ref{tab:results-corner-detection} the results obtained by \numbaselinescdcc \gls*{cnn} models recently proposed for the detection of corners on license plate~images: \smallerlocatenet~\cite{meng2018robust}, \locatenet~\cite{meng2018robust} and \yoohybrid~\cite{yoo2020deep}.
For a fair comparison and considering that the classification of the meters as legible/operational or illegible/faulty is essential in the proposed \gls*{amr} pipeline, we added an output layer (softmax) to each baseline so that they can also perform such classification.
We emphasize that, according to our experiments,  this additional layer does not significantly affect the results obtained in the corner detection task (the normalized mean pixel distance achieved with and without that layer varied slightly in the fourth decimal~place).

\begin{table}[!htb]
\setlength{\tabcolsep}{8pt}
\centering
\caption{\small Comparison of the corner detection results obtained by \gls*{cdcc} and \numbaselinescdcc baselines.
The proposed model presents similar accuracy to \yoohybrid but is twice as fast.
Also, it performs almost as fast as \smallerlocatenet, even though it predicts much more accurate corner~positions.}
\label{tab:results-corner-detection}

\vspace{0.5mm}

\resizebox{0.99\linewidth}{!}{ %
\begin{tabular}{@{}lcccc@{}}
\toprule
\multicolumn{1}{c}{\multirow{2}{*}{Model}} & \multicolumn{1}{c}{\multirow{2}{*}{\gls*{fps}}} & \multicolumn{3}{c}{\begin{tabular}[c]{@{}c@{}}\small Mean pixel distance between the predicted\\\small corners and the ground truth (normalized) \normalsize\end{tabular}} \\ \cmidrule(l){3-5} 
& & \ufpramr & \phantom{ii}\dataset\phantom{ii} & Average   \\ \midrule

\smallerlocatenet~\cite{meng2018robust} & $\mathbf{207}$ & $0.0059$ & $0.0173$ & $0.0116$ \\
\locatenet~\cite{meng2018robust} & $166$ & $0.0031$ & $0.0098$ & $0.0065$ \\
\gls*{cdcc} (Ours) & $\fpscdcc$ & $0.0017$ & $0.0055$ & $0.0036$ \\
\yoohybrid~\cite{yoo2020deep} & $97$ & $\mathbf{0.0016}$ & $\mathbf{0.0046}$ & $\mathbf{0.0031}$ \\\bottomrule
\end{tabular}
} %
\end{table}

As can be seen, \gls*{cdcc} presents the best balance between accuracy and speed among the evaluated models.
More specifically, (i)~\smallerlocatenet is considerably less accurate in predicting the corner positions than the other networks, even though it runs faster; %
(ii)~\gls*{cdcc} is both faster and more accurate in locating the corners than~\locatenet; and
(iii)~\gls*{cdcc} predicts the positions of the corners almost as precisely as \yoohybrid, despite being able to process twice as many~\gls*{fps}. 

\major{Fig.~}\ref{fig:cdcc-net-loss-plot}\major{ shows the evolution of the training and validation losses of CDCC-NET over time (we omitted the losses related to counter classification for better viewing).
As it can be seen, CDCC-NET learns to predict the position of the four corners simultaneously and converged after $14$/$15$~epochs.}

\begin{figure}[!htb]
    \centering
    \includegraphics[width=0.99\linewidth]{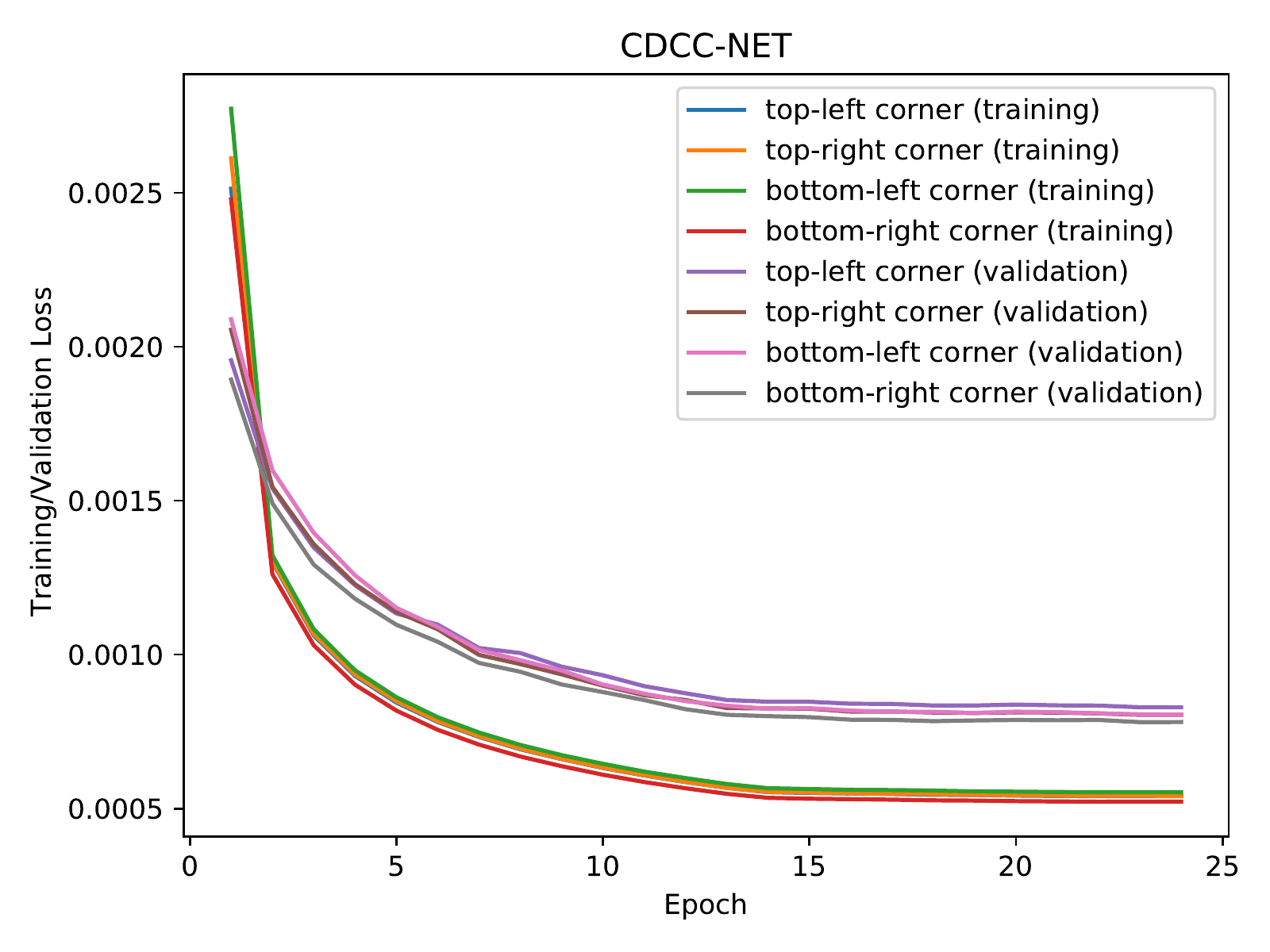}
    
    \vspace{-4mm}
    
    \caption{\small \major{Training and validation losses of CDCC-NET over time. For each corner, the loss plotted is the mean between the $x$ and $y$ coordinates, e.g., the `top-left corner' loss is the mean between the losses of the $x_0$ and $y_0$~tasks.}}
    \label{fig:cdcc-net-loss-plot}
\end{figure}

It should be noted that we evaluated deeper networks in place of \gls*{cdcc} (i.e., with more convolutional layers and/or more filters), however, the end-to-end reading results achieved by the proposed \gls*{amr} system improved only slightly --~not justifying the higher computational cost.
In fact, we carried out an experiment in which we rectified all counter regions using the ground-truth annotations, as if the four corners were detected perfectly on each image, and the results improved less than we expected (from~$\accaverage$\% to~$\accaveragegt$\%), implying that most reading errors made by our \gls*{amr} system were not caused by a poor rectification of the counter region, but by other challenging factors (see Section~\ref{sec:results:overall} for more information and qualitative results). 

Table~\ref{tab:results-counter-classification} shows the results achieved by \gls*{cdcc} in the \textit{counter classification} task, which is significantly less challenging than corner detection.
It is clear that \gls*{cdcc} handles counter classification very well \major{--~in images from both datasets~-- since it correctly filtered out $989$ of the $1{,}000$ images in the Copel-AMR's test set~($\accfaulty$\%)} where it is not possible to perform the meter reading due to occlusions or faulty meters, thereby reducing the overall cost of the proposed system by skipping the counter rectification and recognition tasks in such cases, \major{while correctly accepting $799$ of the $800$ images in the UFPR-AMR's test set ($99.88$\%) and $3{,}990$ of the $4{,}000$ legible/operational images in the Copel-AMR's test set~($99.75$\%).}

\begin{table}[!htb]
\centering
\caption{\small Results achieved by \gls*{cdcc} in the counter classification task.
It is able to filter out \accfaulty\% of the illegible/faulty meters, while correctly accepting \acclegible\% (on average of both datasets) of the legible/operational meters.}
\label{tab:results-counter-classification}

\vspace{0.5mm}

\resizebox{0.725\linewidth}{!}{ %
\begin{tabular}{@{}ccc@{}}
\toprule
\diagbox[trim=l,trim=r]{Dataset}{Class} & Legible/Operational & Illegible/Faulty \\ \midrule

\ufpramr               & $99.88$\%             & -                \\
\dataset                & \major{$99.75$\%}             & $\accfaulty0$\%          \\ \midrule
Average                  & \major{$\acclegible$\%}             & $\accfaulty0$\%          \\ \bottomrule
\end{tabular}
} %
\end{table}

According to Fig.~\ref{fig:results-cdcc}, \gls*{cdcc} is able to successfully predict the four corners of the counter and simultaneously classify it as legible/operational or illegible/faulty, regardless of the meter model and other factors that are common in images acquired in uncontrolled environments such as rotations, reflections and shadows.

\begin{figure}[!htb]
    \centering
    \resizebox{0.995\linewidth}{!}{ %
    \includegraphics[width=0.16\linewidth]{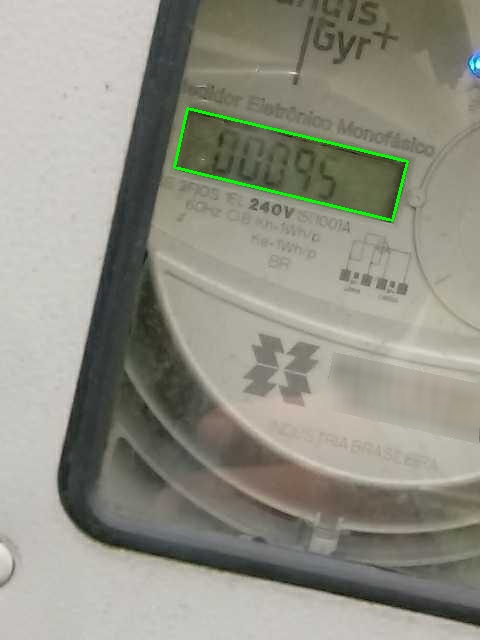}
    \includegraphics[width=0.16\linewidth]{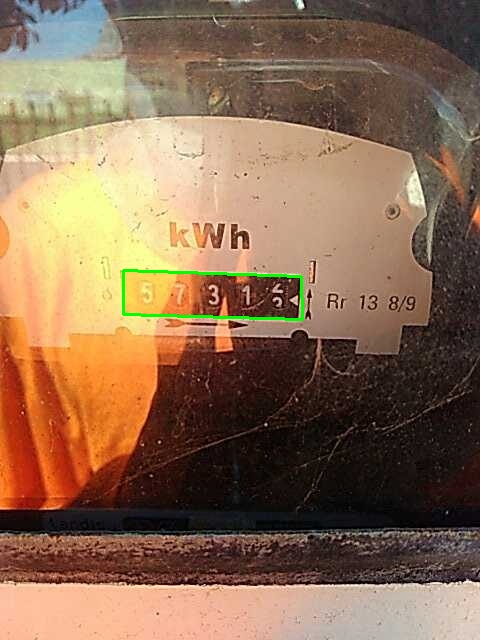}
    \includegraphics[width=0.16\linewidth]{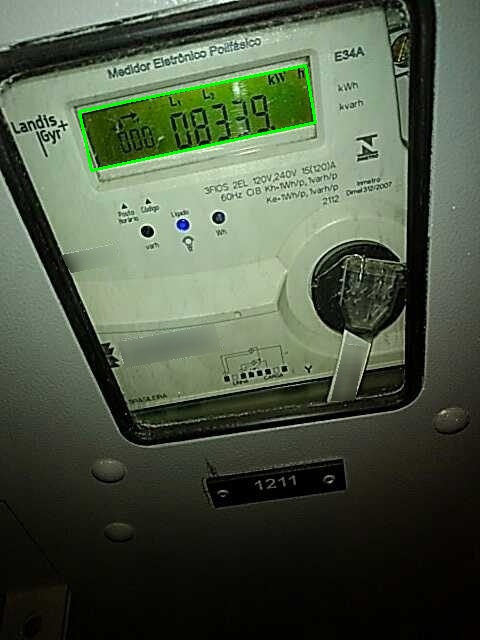}
    \includegraphics[width=0.16\linewidth]{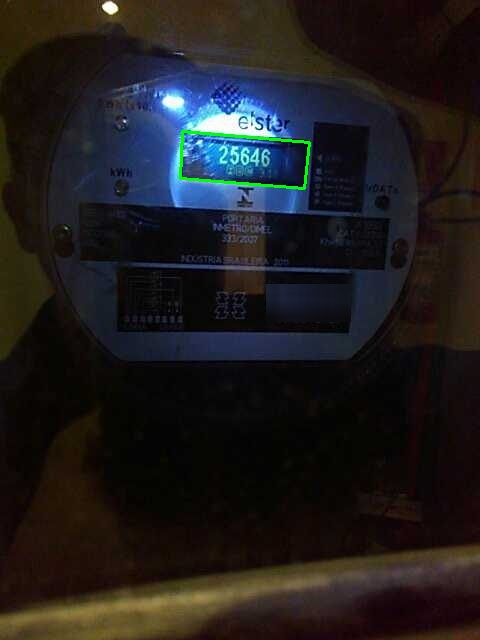}
    \includegraphics[width=0.16\linewidth]{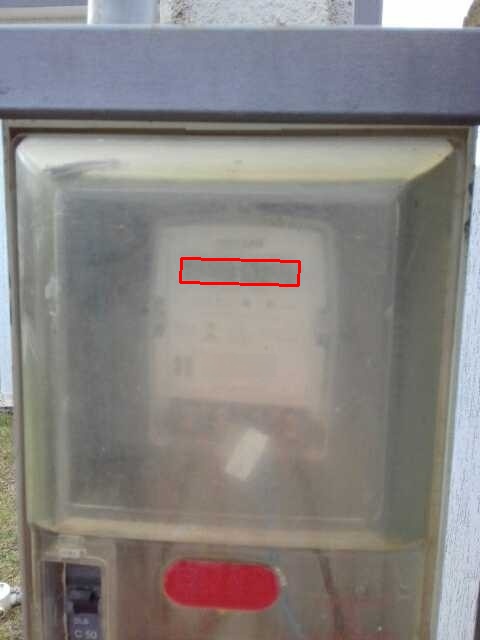}
    \includegraphics[width=0.16\linewidth]{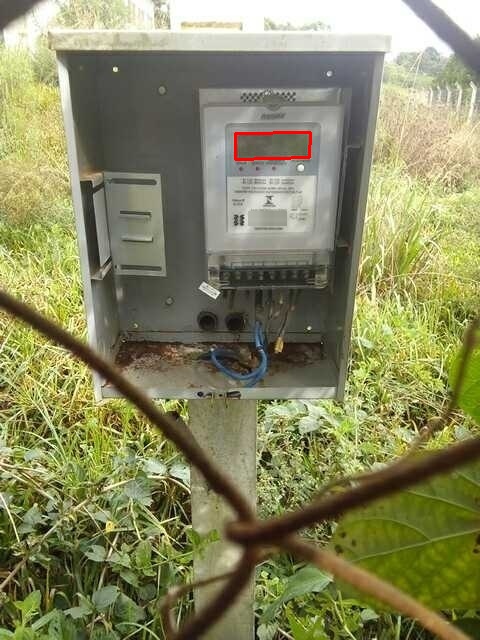}
    } %
    
    \vspace{0.45mm}
    
    \resizebox{0.995\linewidth}{!}{ %
    \includegraphics[width=0.19\linewidth]{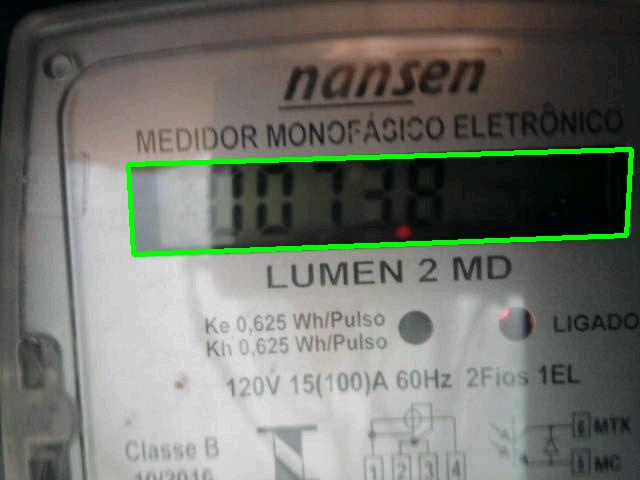}
    \includegraphics[width=0.19\linewidth]{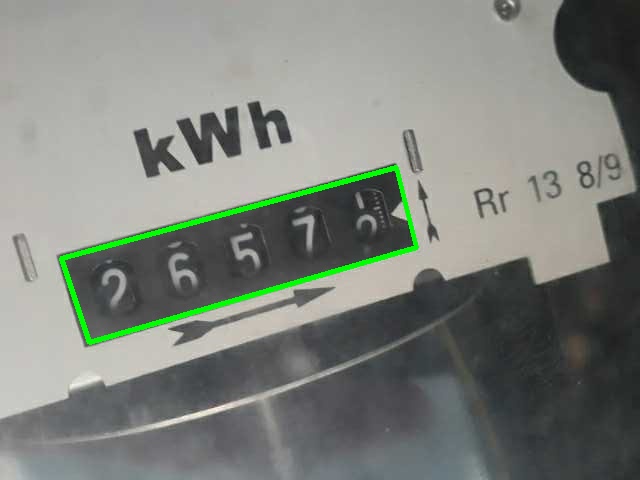}
    \includegraphics[width=0.19\linewidth]{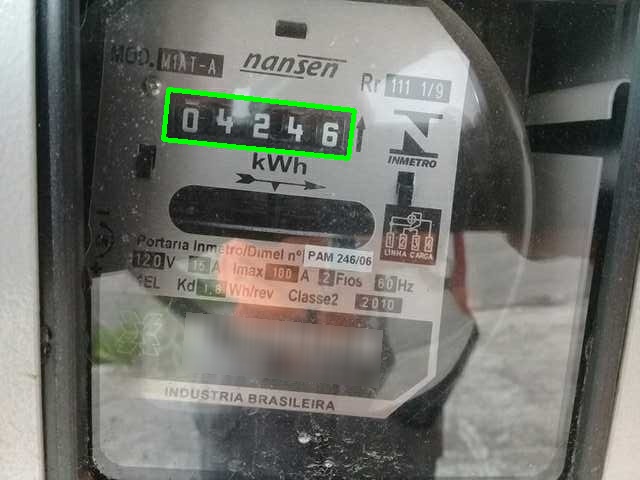}
    \includegraphics[width=0.19\linewidth]{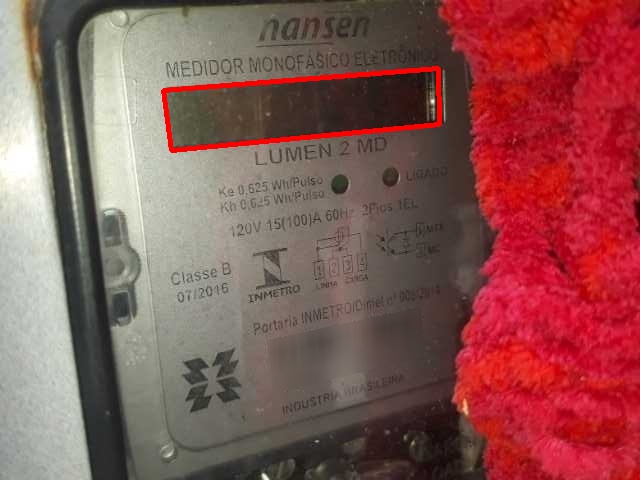}
    \includegraphics[width=0.19\linewidth]{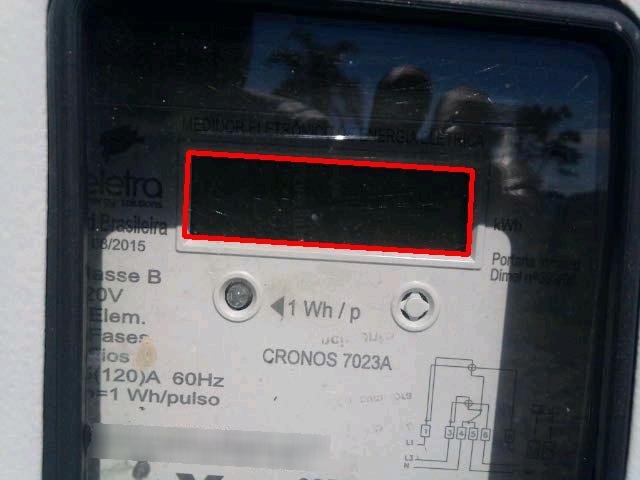}
    } %
    
    \vspace{-1.25mm}
    
    \caption{\small Some qualitative results achieved by \gls*{cdcc} in corner detection and counter classification. For better visualization, we draw a polygon from the predicted corner positions. Counters classified as legible/operational are outlined in green while those classified as illegible/faulty are outlined in~red. 
    }
    \label{fig:results-cdcc}
\end{figure}

\subsubsection{Overall Evaluation (end-to-end)}
\label{sec:results:overall}

In this section, for each dataset, we report the number of correctly recognized~counters divided by the number of legible/operational meters in the test set \major{(legible/operational meters classified as illegible/faulty in the previous stage are considered as a reading error of the proposed method)}.
A correctly recognized counter means that all digits on the counter were correctly recognized, as a single digit recognized incorrectly can result in a large reading/billing error.
As in the previous stage, to enable an accurate analysis regarding the speed/accuracy trade-off of the evaluated methods, we report how many \gls*{fps} each method is capable of processing (in the average of $\nruns$ runs), including the time required to load the respective models and~weights.

\major{Note that our end-to-end system classifies the counter regions as legible/operational or illegible/faulty in the UFPR-AMR dataset as well, even though it does not have images labeled as~illegible/faulty.
This procedure was adopted to better simulate a real-world scenario, where there are images of both legible/operational and illegible/faulty~meters.}


The results obtained by the proposed system and the baselines are shown in Table~\ref{tab:end-to-end-results}.
As can be seen, our system performed the correct reading of $\accufpr$\% of the meters in the \ufpramr's test set and $\accdataset$\% of the legible/operational meters in the \dataset's test set, considerably outperforming all baselines in terms of recognition rate.
It is remarkable that the proposed approach is able to process $\fps$~\gls*{fps} (i.e., it took $\approx\msperimage$~ms to process each image), meeting the efficiency requirements of real-world~applications.

\begin{table*}[!htb]
\centering
\caption{\small Recognition rates obtained by the proposed \gls*{amr} system, a version without counter rectification of our system, and \numbaselines baselines in both datasets used in our experiments. Note that exactly the same images were used for training the proposed methods and the~baselines.}
\label{tab:end-to-end-results}

\vspace{0.5mm}

\resizebox{0.725\linewidth}{!}{ %
\begin{tabular}{@{}lcccc@{}}
\toprule
\multicolumn{1}{c}{\multirow{2}{*}{Approach}} & \multicolumn{1}{c}{\multirow{2}{*}{\gls*{fps}}} & \multicolumn{3}{c}{Recognition Rate} \\ \cmidrule(l){3-5} 
& & \ufpramr & \dataset & Average   \\ \midrule

\detnet + Gómez et al.~\cite{gomez2018cutting} & $66$ & \major{$88.25$\%} & \major{$83.93$\%} & \major{$86.09$\%} \\
\detnet \major{ + CRNN}~\cite{shi2017endtoend} & \major{$69$} & \major{$89.62$\%} & \major{$84.07$\%} & \major{$86.85$\%} \\
Laroca et al.~\cite{laroca2019convolutional} (Multi-task) & $58$ & $89.12$\% & $87.02$\% & $88.07$\% \\
Liao et al.~\cite{liao2019reading} & $61$ & $91.37$\% & $92.25$\% & $91.81$\% \\
\detnet \major{ + Baek et al.}~\cite{baek2019what} \major{(TRBA)} & \major{$31$} & \major{$93.87$\%} & \major{$90.80$\%} & \major{$92.34$\%} \\
\major{YOLOv4 ($416\times416$)}~\cite{bochkovskiy2020yolov4} & \major{$58$} & \major{$91.63$\%} & \major{$93.75$\%} & \major{$92.69$\%} \\
Liao et al.~\cite{liao2019reading} (larger input size) & $42$ & $92.25$\% & $94.27$\% & $93.26$\% \\ %
\detnet + Calefati et al.~\cite{calefati2019reading} & \major{$58$} & $93.87$\% & \major{$95.15$\%} & \major{$94.51$\%} \\
\major{YOLOv4 ($608\times608$)}~\cite{bochkovskiy2020yolov4} & \major{$40$} & \major{$94.00$\%} & \major{$95.33$\%} & \major{$94.67$\%} \\
Laroca et al.~\cite{laroca2019convolutional} (CR-NET) & $41$ & $94.25$\% & $95.20$\% & $94.73$\% \\[2pt] \cdashline{1-5} \\[-5pt]
Ours - unrectified (\detnet + \ocrnet) & $\textbf{\fpsmodified}$ & \major{$\unrectaccufpr$\%} & \major{$\unrectaccdataset$\%} & \major{$\unrectaccaverage$\%} \\
Ours (\detnet + \gls*{cdcc} + \ocrnet) & $\fps$ & $\textbf{\accufpr}$\textbf{\%} & $\textbf{\accdataset}$\textbf{\%} & $\textbf{\accaverage}$\textbf{\%} \\\bottomrule
\end{tabular}
} %
\end{table*}

At first, we were surprised by the fact that the recognition rates reached in the proposed dataset were higher than those obtained in the \ufpramr dataset~\cite{laroca2019convolutional}, where the images were acquired in relatively more controlled conditions.
However, through an inspection of the reading errors made by all methods, we noticed some inconsistencies in the way rotating digits were labeled in the \ufpramr dataset (not always the lowest digit was chosen as the ground truth).
We also observed that the \ufpramr's test set contains some images where it is very difficult to perform the correct reading, even for humans, due to factors such as water vapor, reflections and dirt on the meter glass, as well as the poor positioning of the camera by the person who took the photo, causing the digits to appear only partially in the image even though the counter region appears entirely --~we emphasize that the \ufpramr dataset was collected by one of its authors and not by employees of the service~company, unlike the \dataset dataset.
In fact, in a few cases, it is even difficult to verify if the labeled reading (i.e., the ground truth) is correct.
Although we believe that such images should be rejected by the system due to the great possibility of reading errors (some of these images are shown in Fig.~\ref{fig:samples-challenging-ufpr-amr}), we employed the original labels in our evaluations to enable fair comparisons with other works in the literature, since most authors tend to use the annotations originally provided as part of the dataset.

\begin{figure}[!htb]
    \centering
    \includegraphics[width=0.24\linewidth]{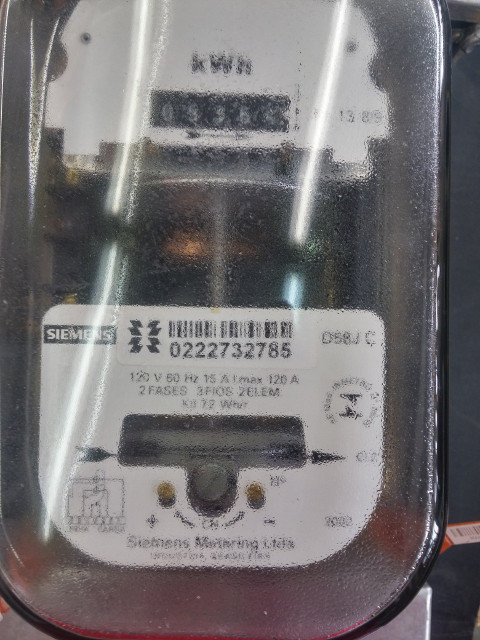} 
    \includegraphics[width=0.24\linewidth]{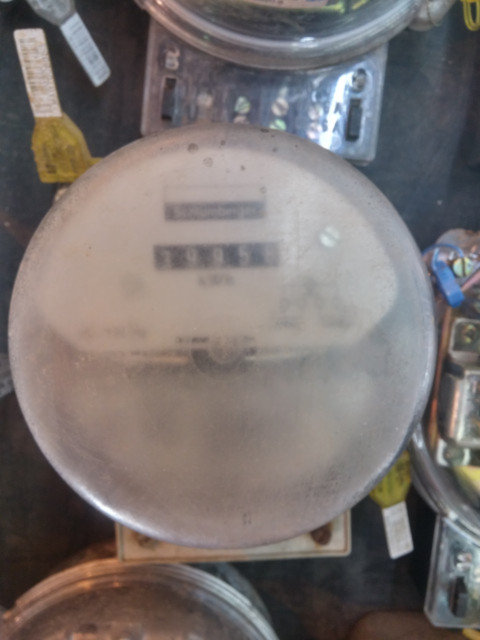} 
    \includegraphics[width=0.24\linewidth]{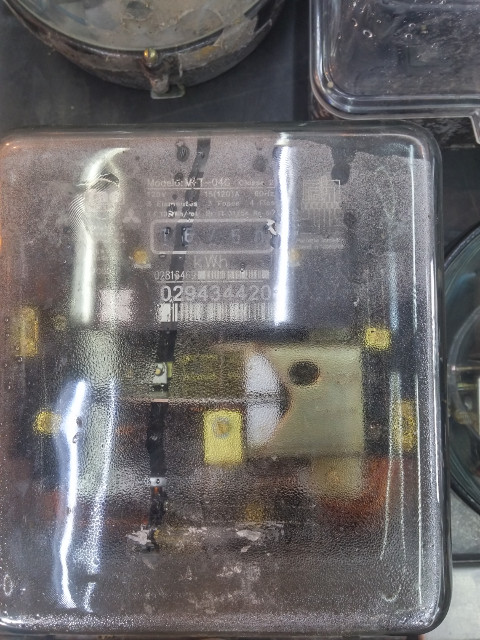} 
    \includegraphics[width=0.24\linewidth]{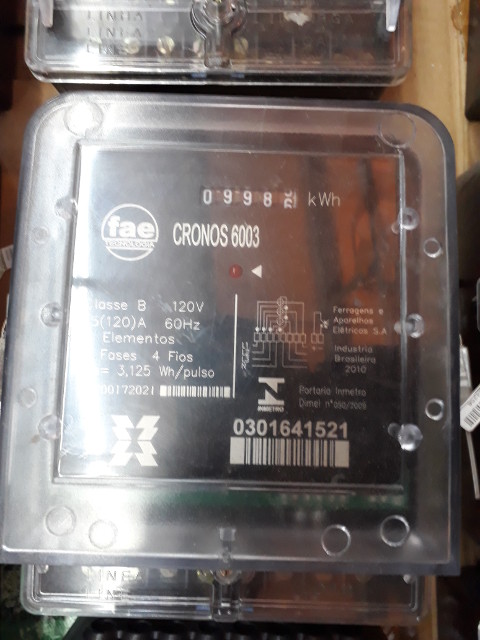}
    
    \vspace{-1.25mm}
    
    \caption{\small Some images from the \ufpramr dataset in which all evaluated methods failed to correctly perform the reading.
    In the first three images, there is dirt or water vapor on the meter glass, while in the fourth image the rightmost digit is labeled as `$6$' and not `$5$', contrary to the protocol normally~adopted.}
    \label{fig:samples-challenging-ufpr-amr}
\end{figure}

To highlight the importance of rectifying the counter region prior to the recognition stage, we included in Table~\ref{tab:end-to-end-results} the results achieved by a modified version of our approach in which the detected counter region is fed directly into \ocrnet \major{ (i.e., without counter rectification)}.
As can be seen, the \textit{corner detection and counter classification} stage is \textbf{essential} for accomplishing outstanding results in unconstrained scenarios, as our system made~$\errorsavoided$\% fewer reading errors  (i.e., $(\accdataset\%-\unrectaccdataset\%)/(100\%-\unrectaccdataset\%)$) \major{in the legible/operational meters} of the \dataset dataset when feeding rectified counters into the recognition~network.


The end-to-end results achieved by the baselines ranged from $\minbaselines$\% to $\maxbaselines$\%.
As some of them were originally evaluated only on private datasets, it is now possible to assess their applicability --~both in terms of speed and accuracy~-- more accurately.
For example, Gómez et al.~\cite{gomez2018cutting} reported a promising recognition rate of $94.17$\% on a proprietary dataset using their segmentation-free network.
Nevertheless, in our comparative assessment, this model reached the lowest recognition rate among the baselines.
In general, as observed in~\cite{laroca2019convolutional}, the recognition models  based on object detectors (e.g., CR-NET and \ocrnet~--~which are based on YOLO~\cite{redmon2016yolo}) performed better than those where counter recognition is done holistically (e.g.,~\cite{shi2017endtoend,gomez2018cutting,baek2019what,calefati2019reading}).

In terms of execution time, we noticed that all methods evaluated are relatively fast, i.e., they are all capable of processing more than $30$ \gls*{fps} on a high-end GPU, especially the version without counter rectification of our system, which is able to process $\fpsmodified$~\gls*{fps}.
\major{As can be seen clearly in Fig.~}\ref{fig:performance-graph}\major{, the proposed system outperformed all baselines in terms of average recognition rate while being relatively fast, and its version without counter rectification is much faster than the baselines that reached similar~results.}

\begin{figure}[!htb]
    \centering
    \includegraphics[width=\linewidth]{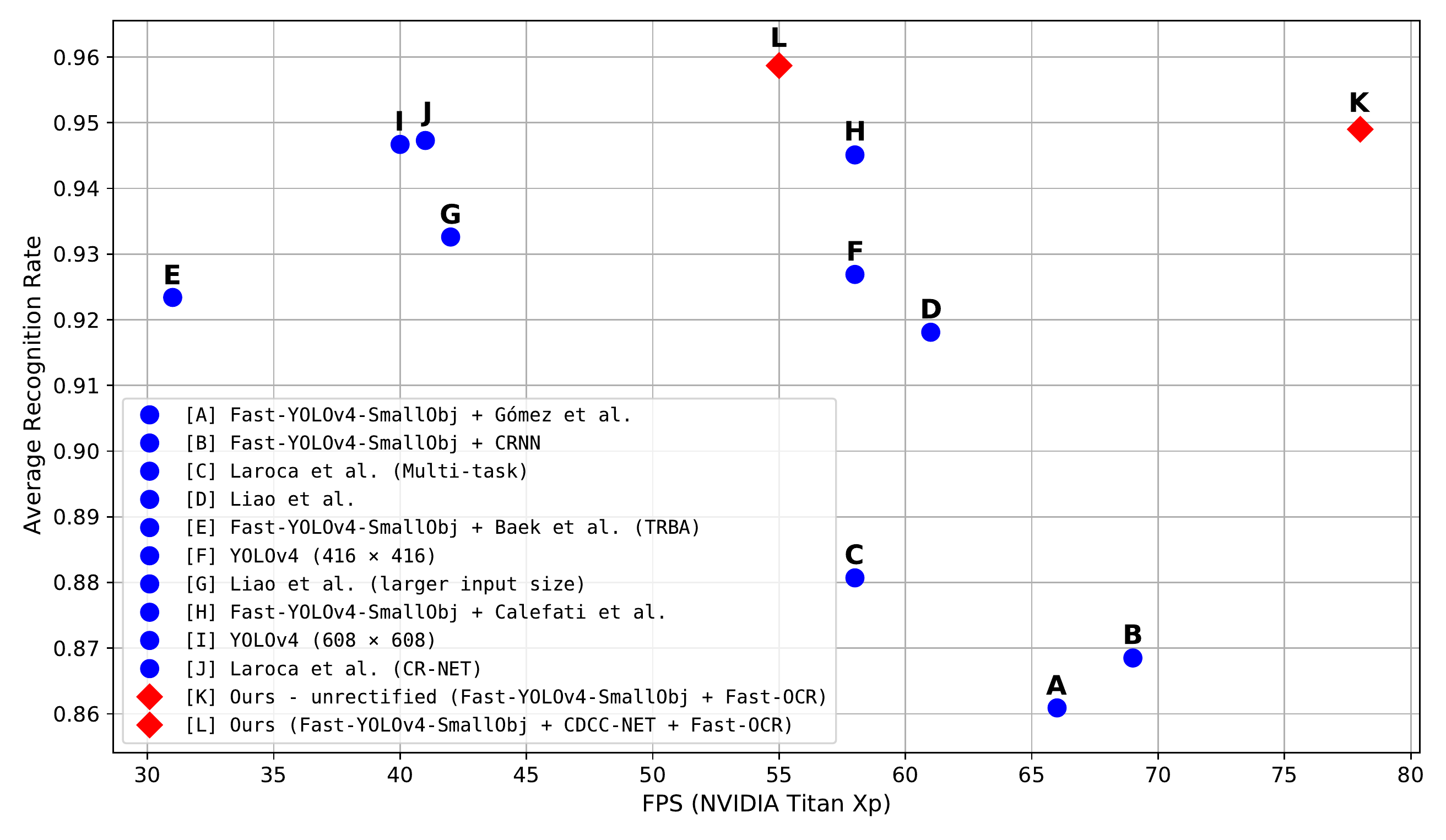}
    
    \vspace{-3mm}
    
    \caption{\small \major{Performance comparison of our AMR system with $\numbaselines$ deep learning-based baselines. 
    For better viewing, both the proposed method~(L) and its version without counter rectification~(K) are depicted as red diamonds, whereas all the baselines~(A-J) are depicted as blue~circles.}}
    \label{fig:performance-graph}
\end{figure}

\major{We highlight that if an AMR system can run at $30$+~FPS on a high-end~GPU, it probably also works well on cheaper hardware (this is relevant to the service company).}
In this sense, for simpler/constrained scenarios, we believe that the proposed AMR system can be employed in low-end setups or even in some mobile phones (taking a few~seconds).



Fig.~\ref{fig:end-to-end-results-correct} shows some meter readings performed correctly by the proposed system.
It is noticeable that our end-to-end system is able to generalize well, being robust to meters of different models and images captured in unconstrained conditions (e.g., with various lighting conditions, reflections, shadows, scale variations, considerable rotations,~etc.).

\begin{figure*}[!htb]
	\centering
 	\captionsetup[subfigure]{labelformat=empty,captionskip=1.25pt,font={footnotesize}} 
	
	%
	\resizebox{0.8\linewidth}{!}{ %
	\subfloat[][\texttt{04241}]{
		\includegraphics[width=0.16\linewidth]{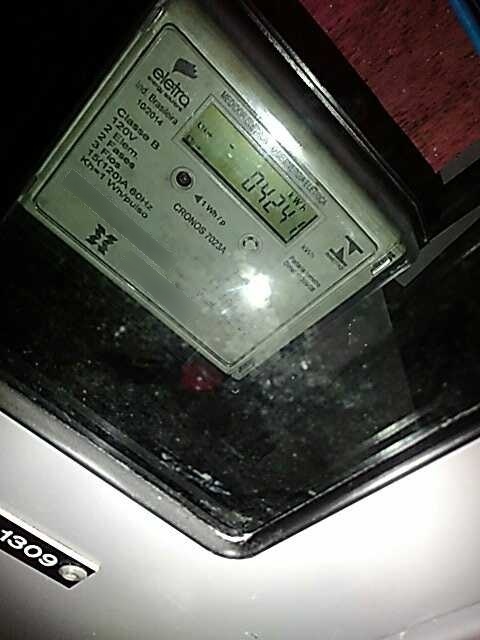}}%
	\subfloat[][\texttt{25464}]{
		\includegraphics[width=0.16\linewidth]{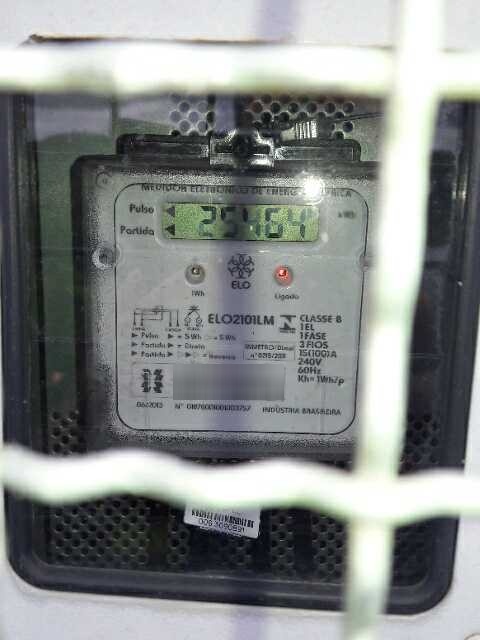}}%
    \subfloat[][\texttt{57599}]{
		\includegraphics[width=0.16\linewidth]{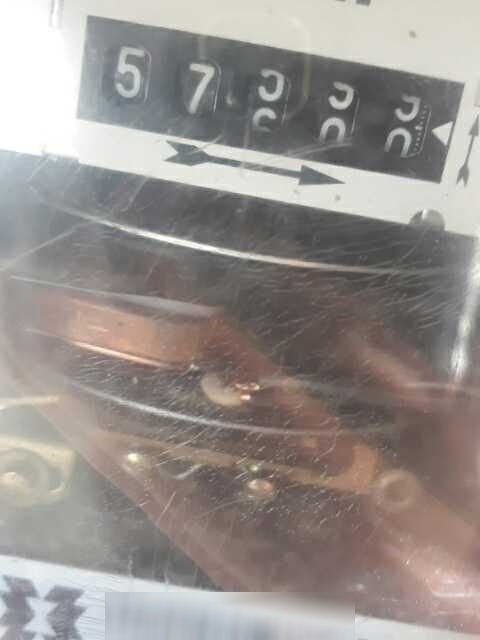}}%
    \subfloat[][\texttt{00841}]{
    \includegraphics[width=0.16\linewidth]{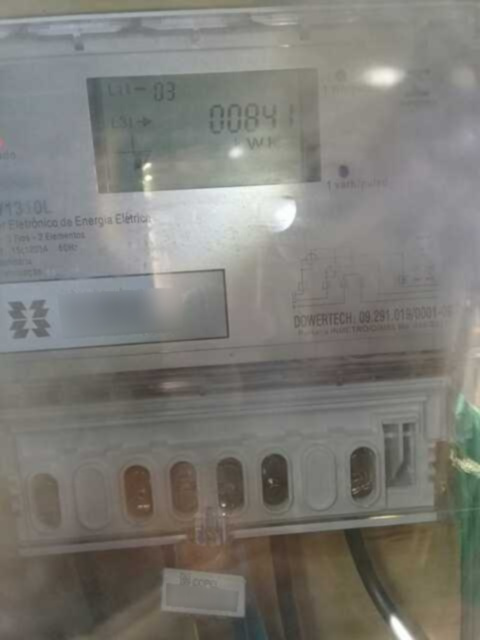}}
	\subfloat[][\texttt{44671}]{
		\includegraphics[width=0.16\linewidth]{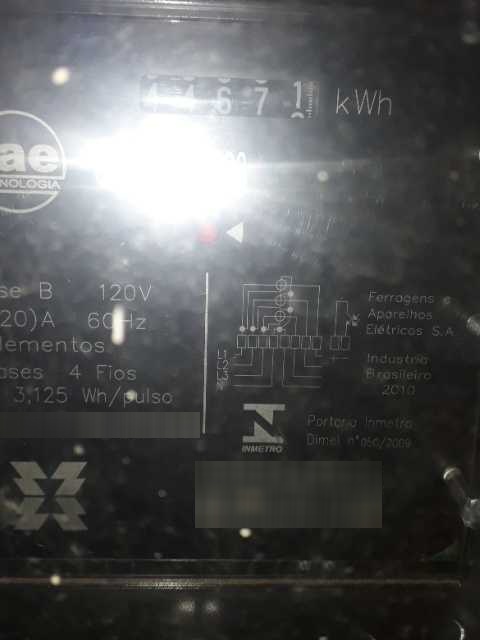}}%
	\subfloat[][\texttt{04730}]{
		\includegraphics[width=0.16\linewidth]{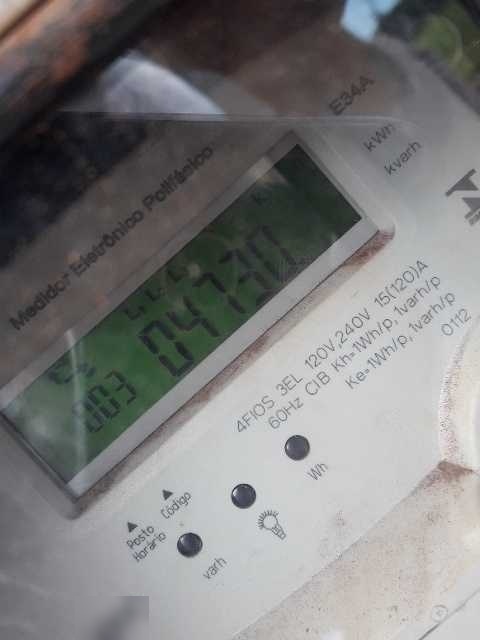}} \hspace{1.25mm}
	} 

	\vspace{1.1mm}
	
	%
	\resizebox{0.8\linewidth}{!}{ %
	\subfloat[][\texttt{25725}]{
		\includegraphics[width=0.19\linewidth]{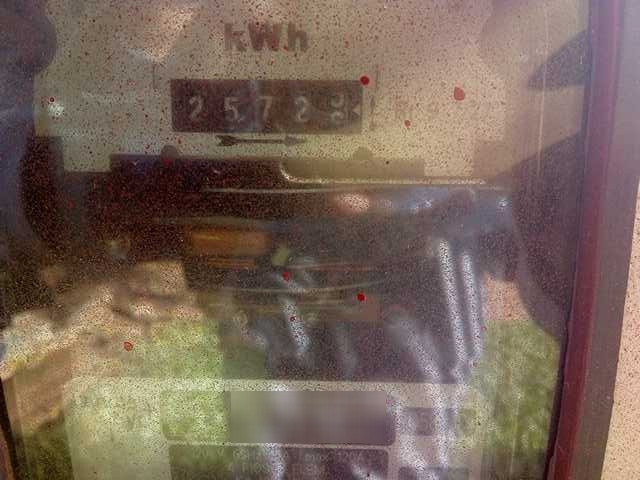}}%
	\subfloat[][\phantom{a}\texttt{10651}]{
		\includegraphics[width=0.19\linewidth]{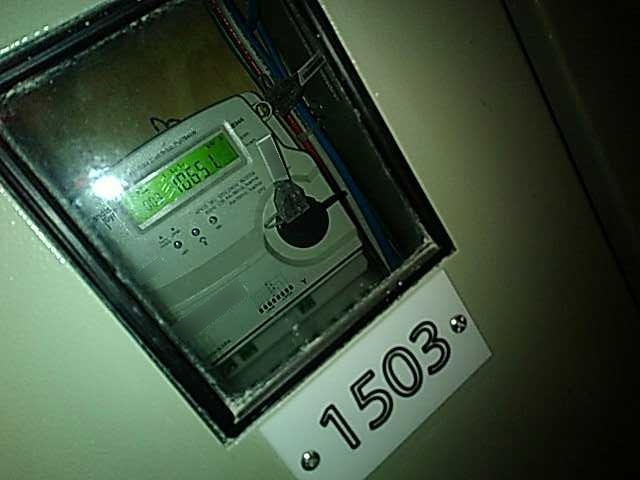}}%
    \subfloat[][\phantom{a}\texttt{06578}]{
		\includegraphics[width=0.19\linewidth]{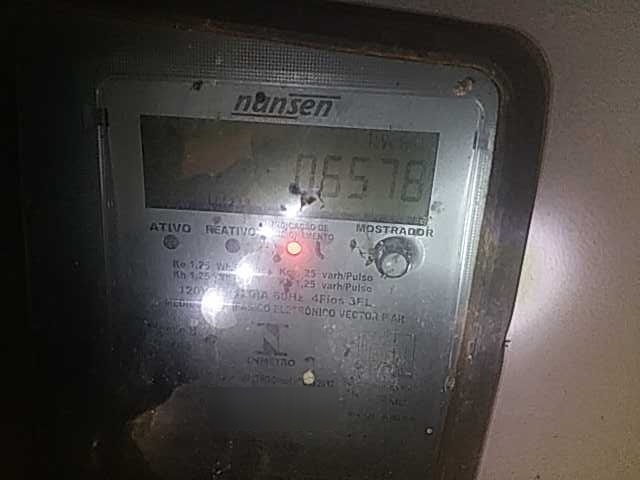}}%
	\subfloat[][\texttt{00353}]{
		\includegraphics[width=0.19\linewidth]{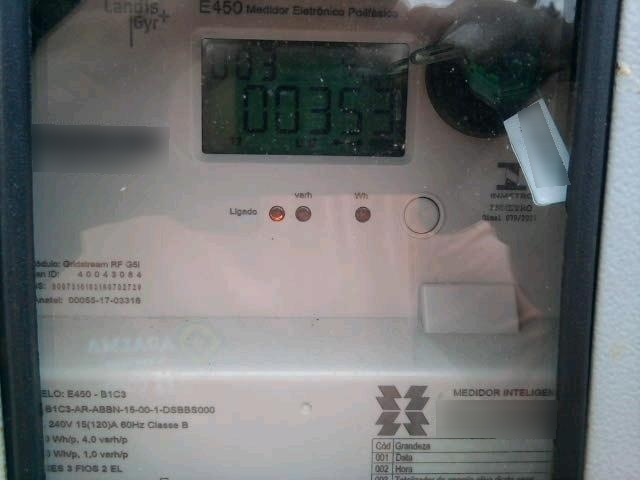}}%
	\subfloat[][\texttt{02059}]{
		\includegraphics[width=0.19\linewidth]{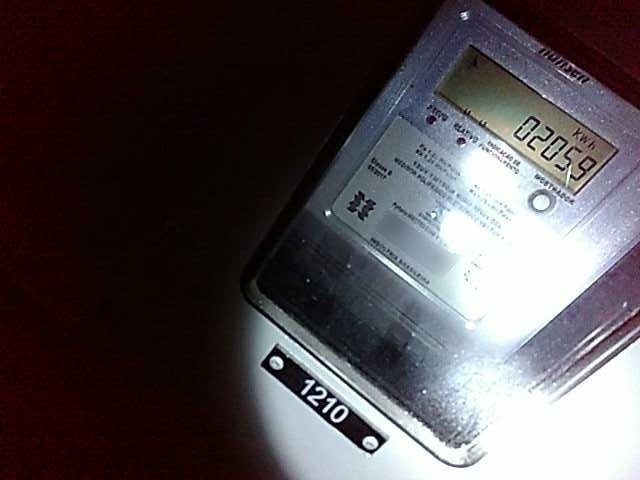}} \hspace{1.25mm}
	} 

	\vspace{1.1mm}
	
	%
	\resizebox{0.8\linewidth}{!}{ %
	\subfloat[][\texttt{07059}]{
		\includegraphics[width=0.16\linewidth]{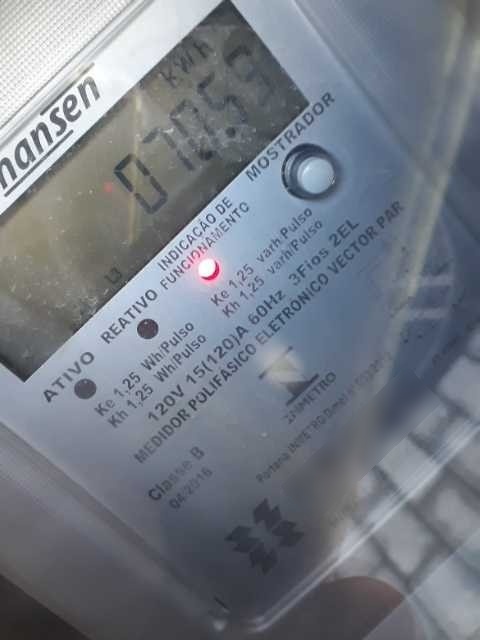}}
	\subfloat[][\texttt{21270}]{
		\includegraphics[width=0.16\linewidth]{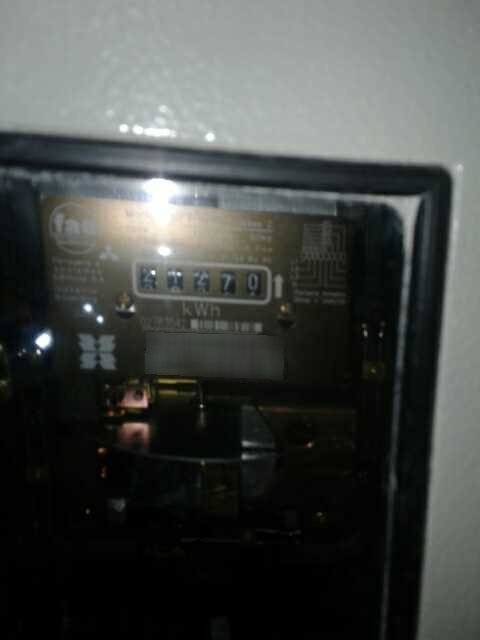}}%
	\subfloat[][\texttt{18101}]{
		\includegraphics[width=0.16\linewidth]{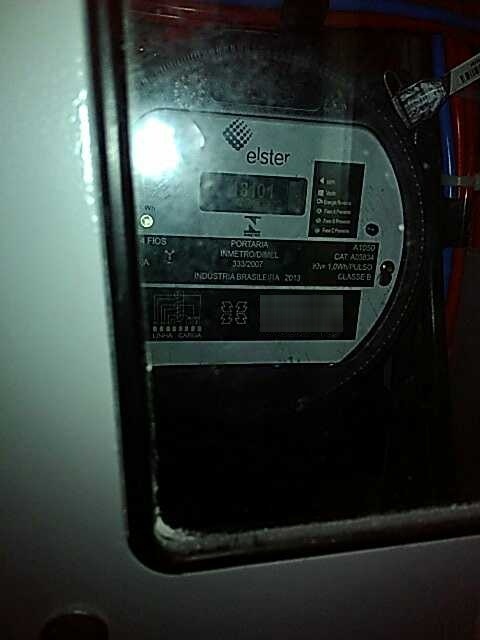}}%
    \subfloat[][\texttt{01710}]{
 		\includegraphics[width=0.16\linewidth]{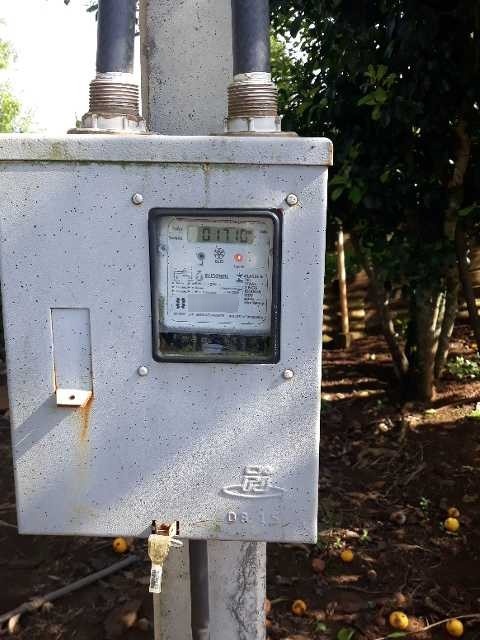}}%
    \subfloat[][\texttt{03953}]{
		\includegraphics[width=0.16\linewidth]{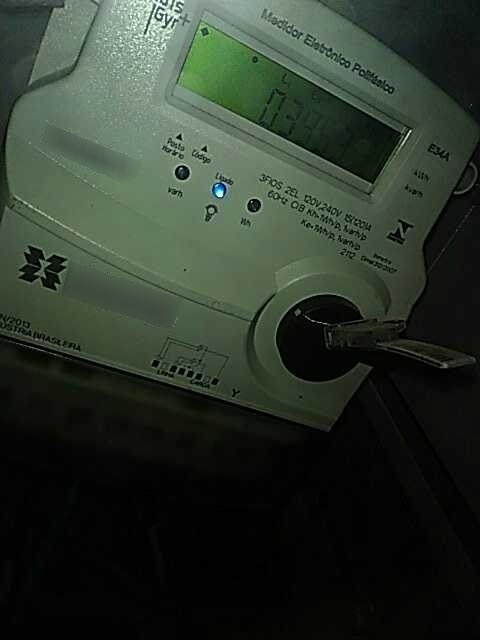}}%
	\subfloat[][\texttt{28382}]{
		\includegraphics[width=0.16\linewidth]{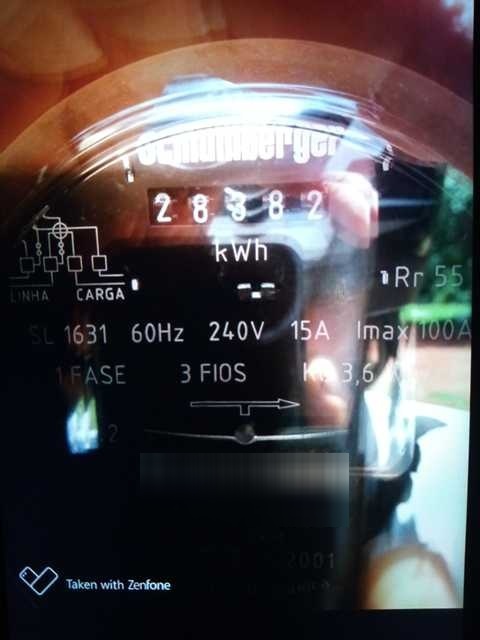}} \hspace{1.25mm}
	} 

	\vspace{-1.1mm}

    \caption{\small Examples of meter readings performed correctly by the proposed system.
    It is remarkable that it performed well in images of meters of different models and captured in unconstrained conditions (e.g., with various lighting conditions, reflections, shadows, scale variations, and considerable~rotations).}
    \label{fig:end-to-end-results-correct}  
\end{figure*}

Some reading errors made by our \gls*{amr} system are shown in Fig.~\ref{fig:end-to-end-results-errors}.
As one may see, they occurred mainly in challenging cases, where one digit becomes very similar to another due to artifacts in the counter region.
Another portion of the errors occurred on rotating digits (also called half digits), which is known to be a major cause of errors in electromechanical~meters~\cite{gao2018automatic,laroca2019convolutional}, even when robust approaches/models are employed for digit/counter~recognition.
It should be noted that (i)~errors in the least significant digits are tolerable, as they do not significantly impact the amount charged to consumers; and (ii) reading errors in the most significant digits can be filtered by the service company through heuristic rules, for example, the reading must be greater than or equal to the reading taken in the previous~month.

\begin{figure*}[!htb]
	\centering
 	\captionsetup[subfigure]{labelformat=empty,captionskip=1.25pt,font={footnotesize},justification=centering} 
	
	\resizebox{0.8\linewidth}{!}{ %
	\subfloat[][\texttt{1874\textcolor{red}{0}}\hspace{\textwidth}\texttt{(18748)}]{
		\includegraphics[width=0.16\linewidth]{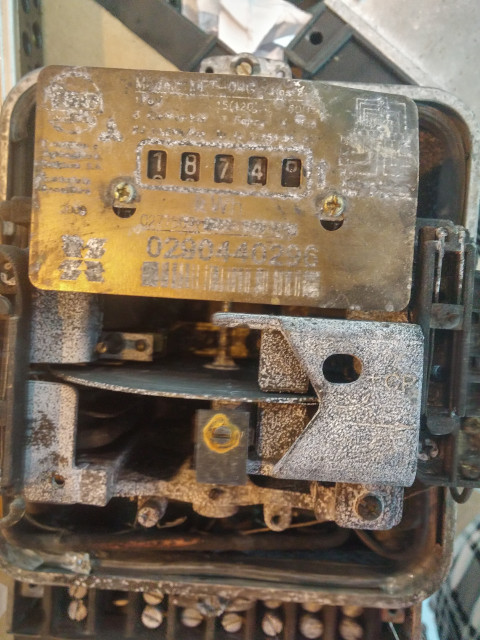}}%
	\subfloat[][\texttt{18\textcolor{red}{3}27}\hspace{\textwidth}\texttt{(18527)}]{
		\includegraphics[width=0.16\linewidth]{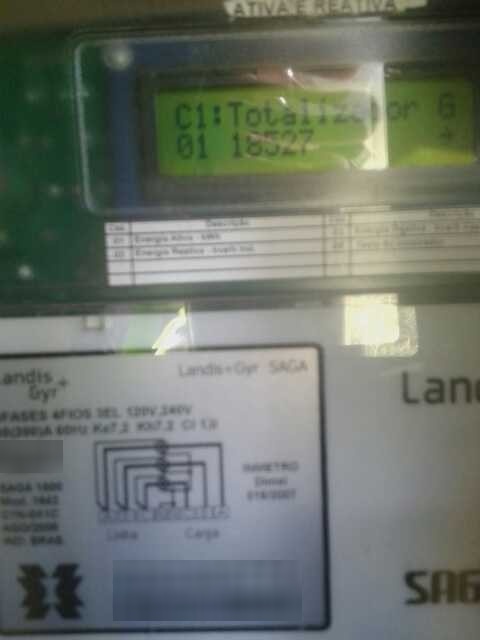}}
    \subfloat[][\texttt{104\textcolor{red}{6}9}\hspace{\textwidth}\texttt{(10459)}]{
		\includegraphics[width=0.16\linewidth]{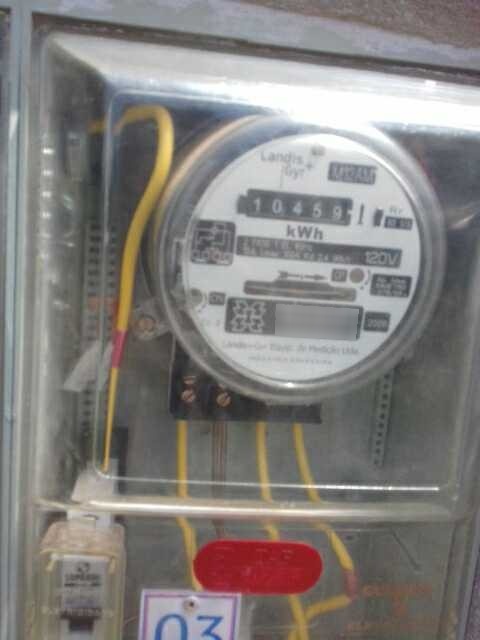}}%
    \subfloat[][\texttt{09\textcolor{red}{3}79}\hspace{\textwidth}\texttt{(09979)}]{
	\includegraphics[width=0.16\linewidth]{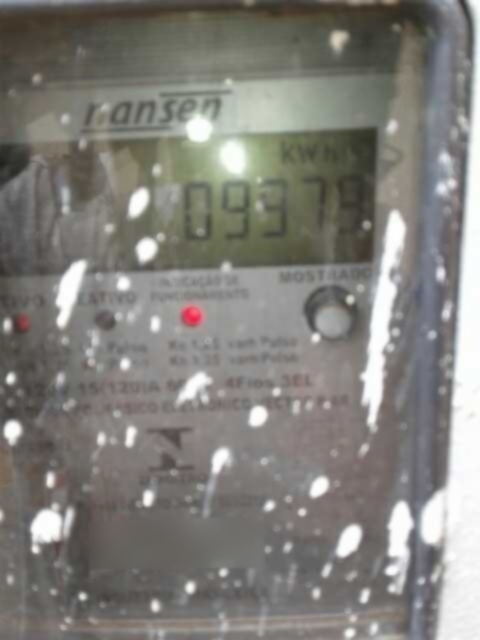}}%
		
	\subfloat[][\texttt{00\textcolor{red}{2}64}\hspace{\textwidth}\texttt{(00864)}]{
		\includegraphics[width=0.16\linewidth]{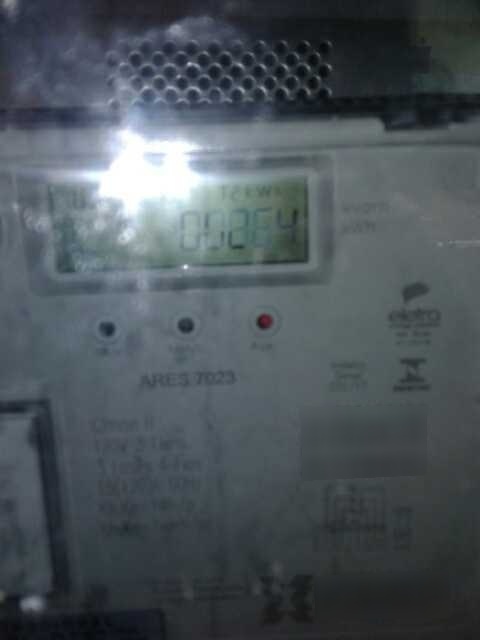}}%
    \subfloat[][\texttt{3412\textcolor{red}{5}}\hspace{\textwidth}\texttt{(34124)}]{
		\includegraphics[width=0.16\linewidth]{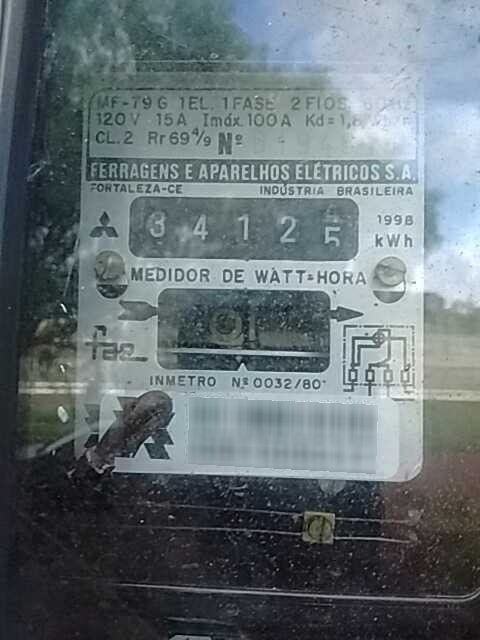}} \hspace{1.25mm}
	} 

	\vspace{1.1mm}
	
	%
	\resizebox{0.8\linewidth}{!}{ %
	\subfloat[][\texttt{0\textcolor{red}{1}815}\hspace{\textwidth}\texttt{(07815)}]{	\includegraphics[width=0.19\linewidth]{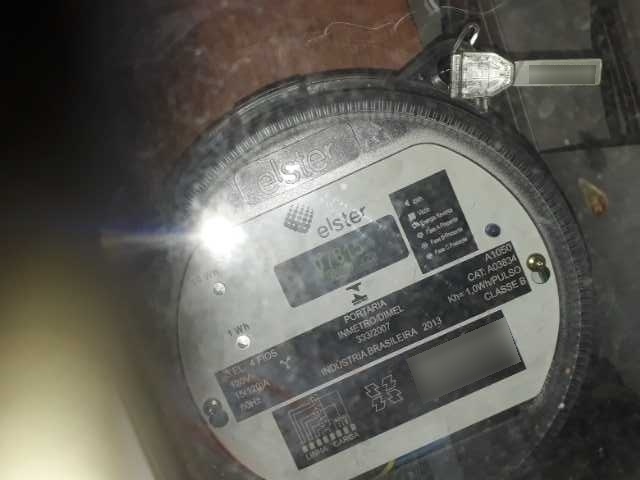}}
	\subfloat[][\texttt{0393\textcolor{red}{1}}\hspace{\textwidth}\texttt{(03937)}]{
	\includegraphics[width=0.19\linewidth]{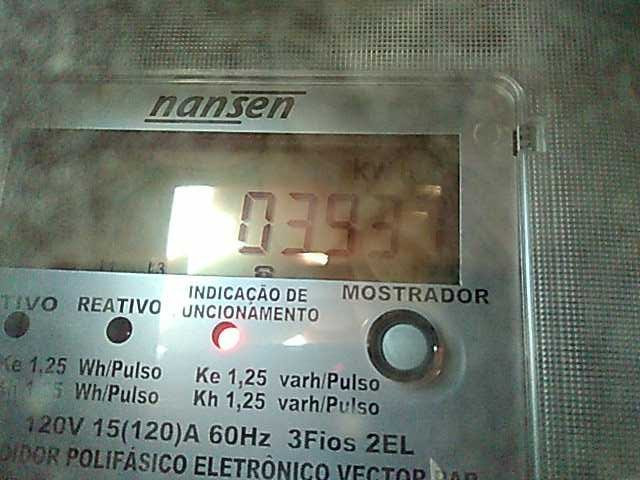}}
    \subfloat[][\texttt{6305\textcolor{red}{4}}\hspace{\textwidth}\texttt{(63059)}]{
	\includegraphics[width=0.19\linewidth]{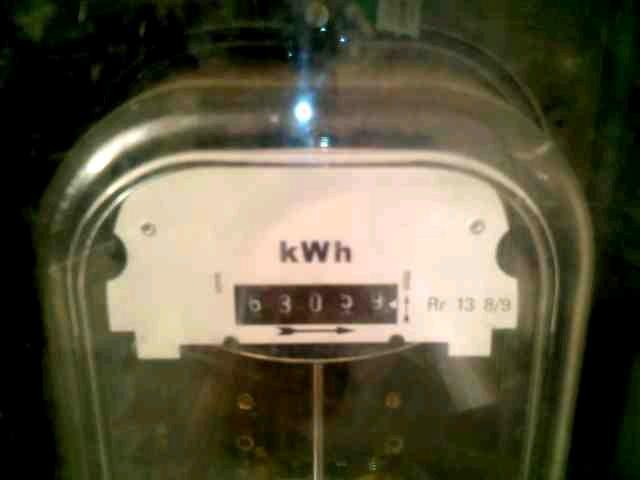}}%
	\subfloat[][\texttt{02\textcolor{red}{9}23}\hspace{\textwidth}\texttt{(02323)}]{
	\includegraphics[width=0.19\linewidth]{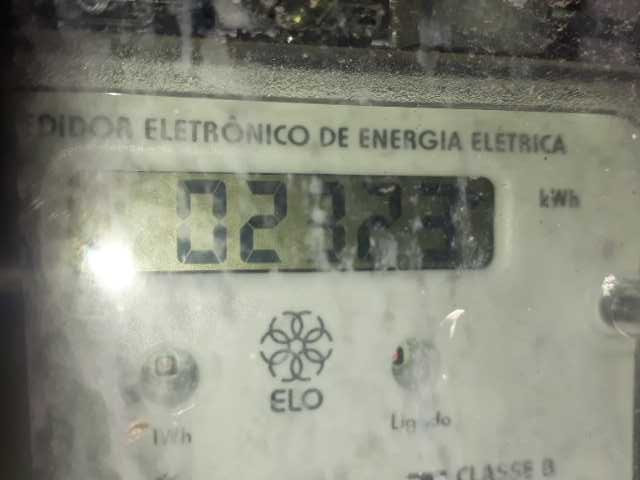}}
    \subfloat[][\texttt{2409\textcolor{red}{1}}\hspace{\textwidth}\texttt{(24097)}]{
	\includegraphics[width=0.19\linewidth]{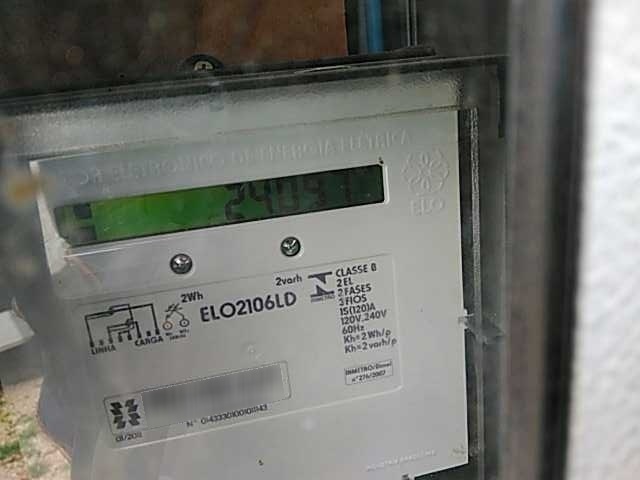}}
	} 

	\vspace{-1.1mm}

    \caption{\small Examples of reading errors made by our system.
	The ground truth is shown in parentheses.
    Observe that most of the errors occurred in challenging cases, where even humans can make mistakes, as one digit becomes very similar to another due to rotating digits or artifacts in the counter~region.}
    \label{fig:end-to-end-results-errors}  
\end{figure*}

\begin{table}[!htb]
\centering
\setlength{\tabcolsep}{7pt}
\caption{\small Recognition rates reached by the proposed \gls*{amr} system when discarding/rejecting the readings returned with lower confidence values by the \ocrnet network. Our system achieves impressive recognition rates (i.e.,~$\ge$~$99$\% on average) when using a confidence threshold that rejects $15$\% of the~images.}
\label{tab:results-with-rejection}

\vspace{0.5mm}

\resizebox{0.975\linewidth}{!}{ %
\begin{tabular}{@{}cccccc@{}}
\toprule
\multirow{2}{*}{Dataset} & \multicolumn{5}{c}{Rejection Rate}            \\ \cmidrule(l){2-6} 
                                          & $0$\%     & $5$\%    & $10$\%    & $15$\%    & $20$\%    \\ \midrule

\ufpramr & $\accufpr$\% & $97.63$\% & $98.47$\% & $98.82$\% & $99.22$\% \\
\dataset                                 & $\accdataset$\% & $98.61$\% & $99.00$\% & $99.24$\% & $99.44$\% \\ \midrule
Average                                   & $\accaverage$\% & $98.12$\% & $98.74$\% & $99.03$\% & $99.33$\% \\ \bottomrule \\[-12.5pt]
\end{tabular}
}
\end{table}

Finally, considering that very few reading errors are tolerated by \gls*{copel}~\cite{copel} and other service companies, we present in Table~\ref{tab:results-with-rejection} the end-to-end results achieved by the proposed system when discarding/rejecting the readings returned with lower confidence values by the \ocrnet network (in practice, in a mobile application, the employee would have to capture another image).
It is noteworthy that our \gls*{amr} system achieved an average recognition rate above~$98$\% by rejecting only $5$\%~of the meter readings.
Moreover, recognition rates above $99$\%, which are acceptable to service companies, are achieved by setting a confidence threshold that rejects $15$\% of the~images.
In this way, we consider that the proposed approach can be reliably employed on real-world~applications.
\section{Conclusions and Future Work}
\label{sec:conclusions}

In this work, we presented an end-to-end, robust and efficient approach for \gls*{amr} that achieves state-of-the-art results in two public datasets while being able to significantly reduce the number of images that are sent to human review by filtering out images in which it is not possible to perform the meter reading due to occlusions or faulty meters.
Our main contribution is the insertion of a new stage in the \gls*{amr} pipeline, called \textit{corner detection and counter classification}, which enables the counter region to be rectified prior to the recognition stage.
As the proposed system made~$\errorsavoided$\% fewer reading errors \major{in the legible/operational meters} of the \dataset dataset when feeding rectified counters into the recognition~network, we consider this strategy (corner detection + counter rectification) essential for accomplishing outstanding results in unconstrained~scenarios.

Our \gls*{amr} system, which presents three new networks operating in a cascaded mode, performed the correct reading of \accufpr\% and \accdataset\% of the meters in the \ufpramr and \dataset test sets (outperforming all baselines), respectively, while being able to process \fps~\gls*{fps} on a high-end GPU.
It is notable that the proposed approach achieves impressive end-to-end recognition rates (i.e.,~$\ge$~99\%) when discarding/rejecting the readings made with lower confidence~values, which is of paramount importance to the service companies since very few reading errors are tolerated in real-world applications due to the fact that a single digit recognized incorrectly can result in a large reading/billing~error.

We also introduced a publicly available dataset for \gls*{amr} with $\numtotal$ fully-annotated images acquired on real-world scenarios by the service company's employees themselves, including $\numfaulty$ images of faulty meters or cases where the meter reading is illegible due to factors such as shadows and occlusions.
The proposed dataset has six times more images and contains a larger variety in different aspects than the largest dataset found in the literature for the evaluation of end-to-end \gls*{amr}~methods.
It also contains a well-defined evaluation protocol to assist the development of new approaches and the fair comparison among published~works.

As future work, we plan to design a methodology for the simultaneous detection of the counter region and its corners, aiming to perform counter rectification with an even better speed/accuracy~trade-off.
We also intend to explore the meter’s model/type in the \gls*{amr} pipeline and investigate in depth the cases where the counter has rotating digits, considering that this is a major cause of reading errors in electromechanical~meters.
\major{Finally, we want to carry out an extensive assessment with various approaches/models for detecting and recognizing general scene text to compare how robust and efficient they are for the specific AMR scenario.}
\section*{Acknowledgments}

This work was supported in part by the Coordination for the Improvement of Higher Education Personnel~(CAPES) (Grant 88887.516264/2020-00 and Social Demand Program), and in part by the National Council for Scientific and Technological Development~(CNPq) (Grant~308879/2020-1).
We gratefully acknowledge the support of NVIDIA Corporation with the donation of the GPUs used for this~research.
We also thank the \acrfull*{copel}, in particular the manager of the reading division Dihon Pereira Brandão, for providing the images for the creation of the \dataset~dataset.
%

\scriptsize
\balance
\setlength{\bibsep}{3pt}
\bibliographystyle{IEEEtran}
\bibliography{bibtex}


\end{document}